\documentclass{article} % For LaTeX2e
\usepackage[final]{colm2026_conference}

\usepackage{microtype}
\usepackage{hyperref}
\usepackage{url}            % simple URL typesetting
\usepackage{booktabs} % for professional tables
\usepackage{makecell}
\usepackage{lineno}

\definecolor{darkblue}{rgb}{0, 0, 0.5}
\hypersetup{colorlinks=true, citecolor=darkblue, linkcolor=darkblue, urlcolor=darkblue}

\usepackage{graphicx}
\usepackage{subfigure}
\usepackage{amsfonts}       % blackboard math symbols
\usepackage{nicefrac}       % compact symbols for 1/2, etc.
\usepackage{xcolor}         % colors
\usepackage{colortbl}       % for \rowcolor in tables
\usepackage{adjustbox}
\usepackage{amsthm}
\usepackage{epstopdf}
\usepackage{xcolor}
\usepackage{mathtools}
\usepackage{mathrsfs}
\usepackage{graphicx}
\usepackage{subfigure}
\usepackage{paralist}
\usepackage{bm,bbm,amsmath}
\usepackage{wrapfig}
\usepackage[ruled,vlined]{algorithm2e}

\usepackage{forest}
\usepackage{xspace}
\usepackage{multirow}
\usepackage{url}
\usepackage{bm}
\usepackage{diagbox}
\usepackage{float}
\usepackage{placeins}
\allowdisplaybreaks

\newcounter{romancounter}
\newcommand{\rom}[1]{\setcounter{romancounter}{#1}\textit{(\roman{romancounter})}}

\usepackage{amsmath}
\usepackage{amssymb}
\usepackage{mathtools}
\usepackage{amsthm}

\usepackage[capitalize,noabbrev]{cleveref}

\usepackage[textsize=tiny]{todonotes}

\usepackage{tikz}
\usepackage{tikz-3dplot}

\usetikzlibrary{arrows.meta,positioning,calc,backgrounds,shadows.blur,shapes.geometric}

\usepackage{amsmath}
\usepackage{amssymb}
\usepackage{mathtools}
\usepackage{amsthm}
\usepackage{dsfont}
\usepackage{lipsum}
\usepackage{bm,bbm}
\usepackage{bbding}
\usepackage{enumitem}
\usepackage{natbib}%
\usepackage{xspace}
\usepackage[utf8]{inputenc} 
\usepackage[T1]{fontenc}    
\usepackage{hyperref}       
\usepackage{url}            
\usepackage{amsfonts}       
\usepackage{nicefrac}       
\usepackage{microtype} 
\usepackage{multirow}
\usepackage{graphicx}
\usepackage{subfigure}
\usepackage[font=small,labelfont=bf]{caption}
\usepackage{subfloat}
\usepackage{float}
\usepackage{accents}
\usepackage{hhline}
\usepackage{wrapfig}
\usepackage{booktabs}
\usepackage{verbatim}
\usepackage{setspace}
\usepackage{tcolorbox}
\definecolor{mydarkblue}{rgb}{0,0.08,0.45}

\hypersetup{
    colorlinks,
    breaklinks,
    linkcolor = blue,
    citecolor = blue,
    urlcolor  = blue,
}
\def \x {\mathbf{x}}

\def \z {\mathbf{z}}

\def \w {\mathbf{w}}

\def \R {\mathbb{R}}

\def \v {\mathbf{v}}

\def \h {\mathbf{h}}

\def \E {\mathbb{E}}

\def \KL {\textnormal{KL}}

\makeatletter
\let\save@mathaccent\mathaccent
\newcommand*\if@single[3]{%
  \setbox0\hbox{${\mathaccent"0362{#1}}^H$}%
  \setbox2\hbox{${\mathaccent"0362{\kern0pt#1}}^H$}%
  \ifdim\ht0=\ht2 #3\else #2\fi
  }
\newcommand*\rel@kern[1]{\kern#1\dimexpr\macc@kerna}
\newcommand*\widebar[1]{\@ifnextchar^{{\wide@bar{#1}{0}}}{\wide@bar{#1}{1}}}
\newcommand*\wide@bar[2]{\if@single{#1}{\wide@bar@{#1}{#2}{1}}{\wide@bar@{#1}{#2}{2}}}
\newcommand*\wide@bar@[3]{%
  \begingroup
  \def\mathaccent##1##2{%
    \let\mathaccent\save@mathaccent
    \if#32 \let\macc@nucleus\first@char \fi
    \setbox\z@\hbox{$\macc@style{\macc@nucleus}_{}$}%
    \setbox\tw@\hbox{$\macc@style{\macc@nucleus}{}_{}$}%
    \dimen@\wd\tw@
    \advance\dimen@-\wd\z@
    \divide\dimen@ 3
    \@tempdima\wd\tw@
    \advance\@tempdima-\scriptspace
    \divide\@tempdima 10
    \advance\dimen@-\@tempdima
    \ifdim\dimen@>\z@ \dimen@0pt\fi
    \rel@kern{0.6}\kern-\dimen@
    \if#31
      \overline{\rel@kern{-0.6}\kern\dimen@\macc@nucleus\rel@kern{0.4}\kern\dimen@}%
      \advance\dimen@0.4\dimexpr\macc@kerna
      \let\final@kern#2%
      \ifdim\dimen@<\z@ \let\final@kern1\fi
      \if\final@kern1 \kern-\dimen@\fi
    \else
      \overline{\rel@kern{-0.6}\kern\dimen@#1}%
    \fi
  }%
  \macc@depth\@ne
  \let\math@bgroup\@empty \let\math@egroup\macc@set@skewchar
  \mathsurround\z@ \frozen@everymath{\mathgroup\macc@group\relax}%
  \macc@set@skewchar\relax
  \let\mathaccentV\macc@nested@a
  \if#31
    \macc@nested@a\relax111{#1}%
  \else
    \def\gobble@till@marker##1\endmarker{}%
    \futurelet\first@char\gobble@till@marker#1\endmarker
    \ifcat\noexpand\first@char A\else
      \def\first@char{}%
    \macc@nested@a\relax111{\first@char}%
  \fi
  \endgroup
}
\DeclareRobustCommand\onedot{\futurelet\@let@token\@onedot}
\def\@onedot{\ifx\@let@token.\else.\null\fi\xspace}

\makeatother

\definecolor{DSgray}{cmyk}{0,1,0,0}
\definecolor{DSblue}{cmyk}{1,0,0,0}

\definecolor{myblue}{RGB}{68,114,196}

\theoremstyle{definition}

\newtheorem{myRemark}{Remark}
\usepackage{appendix}

\def \epsilon {\varepsilon}

\usepackage[textsize=tiny]{todonotes}
\usepackage{prettyref}

\newcommand{\savehyperref}[2]{\texorpdfstring{\hyperref[#1]{#2}}{#2}}
\newrefformat{eq}{\savehyperref{#1}{Eq.~(\ref*{#1})}}
\newrefformat{eqn}{\savehyperref{#1}{Equation~\ref*{#1}}}
\newrefformat{lem}{\savehyperref{#1}{Lemma~\ref*{#1}}}
\newrefformat{lemma}{\savehyperref{#1}{Lemma~\ref*{#1}}}
\newrefformat{def}{\savehyperref{#1}{Definition~\ref*{#1}}}
\newrefformat{line}{\savehyperref{#1}{Line~\ref*{#1}}}
\newrefformat{thm}{\savehyperref{#1}{Theorem~\ref*{#1}}}
\newrefformat{corr}{\savehyperref{#1}{Corollary~\ref*{#1}}}
\newrefformat{cor}{\savehyperref{#1}{Corollary~\ref*{#1}}}
\newrefformat{sec}{\savehyperref{#1}{Section~\ref*{#1}}}
\newrefformat{app}{\savehyperref{#1}{Appendix~\ref*{#1}}}
\newrefformat{appendix}{\savehyperref{#1}{Appendix~\ref*{#1}}}
\newrefformat{assum}{\savehyperref{#1}{Assumption~\ref*{#1}}}
\newrefformat{ex}{\savehyperref{#1}{Example~\ref*{#1}}}
\newrefformat{fig}{\savehyperref{#1}{Figure~\ref*{#1}}}
\newrefformat{alg}{\savehyperref{#1}{Algorithm~\ref*{#1}}}
\newrefformat{rem}{\savehyperref{#1}{Remark~\ref*{#1}}}
\newrefformat{conj}{\savehyperref{#1}{Conjecture~\ref*{#1}}}
\newrefformat{prop}{\savehyperref{#1}{Proposition~\ref*{#1}}}
\newrefformat{proto}{\savehyperref{#1}{Protocol~\ref*{#1}}}
\newrefformat{prob}{\savehyperref{#1}{Problem~\ref*{#1}}}
\newrefformat{claim}{\savehyperref{#1}{Claim~\ref*{#1}}}
\newrefformat{que}{\savehyperref{#1}{Question~\ref*{#1}}}
\newrefformat{op}{\savehyperref{#1}{Open Problem~\ref*{#1}}}
\newrefformat{fn}{\savehyperref{#1}{Footnote~\ref*{#1}}}
\allowdisplaybreaks

\title{\textsc{AdaFlash}: Adaptive Speculative Decoding via On-Policy Distilled Diffusion Drafters}

\author{%
\textbf{Yu-Yang Qian}\textnormal{\textsuperscript{1,2,*}} ~\;
\textbf{Hao-Cong Wu}\textnormal{\textsuperscript{1,2,*}}  ~\;
\textbf{Chen Chen}\textnormal{\textsuperscript{3}}      ~\; 
\textbf{Jiacheng Sun}\textnormal{\textsuperscript{3}}      ~\;
\textbf{Zhenhua Dong}\textnormal{\textsuperscript{3}}      ~\;\\
\textbf{Peng Zhao}\textnormal{\textsuperscript{1,2,$\dagger$}}  ~\;
\textbf{Zhi-Hua Zhou}\textnormal{\textsuperscript{1,2}} \\
\textsuperscript{1}State Key Laboratory for Novel Software Technology, Nanjing University \\
\textsuperscript{2}School of Artificial Intelligence, Nanjing University ~~~
\textsuperscript{3}Huawei Foundation Model Dept \\[4pt]
}

\begin{document}

\ifcolmsubmission
\linenumbers
\fi

\maketitle
\renewcommand{\thefootnote}{*}
\footnotetext{Equal Contribution. \quad $^\dagger$Corresponding Author: Peng Zhao <zhaop@lamda.nju.edu.cn>.}
\renewcommand{\thefootnote}{\arabic{footnote}}

\begin{abstract}
    Speculative decoding, in which a lightweight draft model first generates a draft sequence that is then verified by the target model, has become a prevalent paradigm for accelerating large language model inference. Recent work such as DFlash further boosts drafting efficiency by leveraging diffusion drafters, whose parallel denoising mechanism enables draft generation in a single forward pass. In this work, we uncover a \emph{central pitfall} of diffusion drafters: the \emph{global dependency}, arising from bidirectional attention and KV injection in diffusion drafters, is a double-edged sword. On one hand, it endows the model with parallel generation and global contextual modeling capabilities; on the other hand, this global dependency introduces \emph{high variance} at both the domain-level and the token-level: acceptance rates fluctuate substantially across different domains, and the acceptance probability varies across draft positions. To tackle this issue, we propose the \textsc{AdaFlash} framework, comprising two components: (i) an \emph{on-policy distillation} (OPD) algorithm with reverse-KL divergence tailored for diffusion drafters, which delivers stable convergence and continuously adapts the drafter to the deployment distribution; and (ii) an \emph{adaptive length head} that dynamically adjusts the candidate length on the fly, substantially lowering the verification cost of the target model. Experiments demonstrate that \textsc{AdaFlash} consistently improves the speedup during deployment, with especially significant gains under high-concurrency, achieving up to 66\% higher average throughput than previous state-of-the-art methods. Our code is available at \url{https://github.com/ZinYY/AdaFlash}.
\end{abstract}

\section{Introduction}
\label{sec:introduction}

Large language models~(LLMs) have demonstrated impressive capabilities across a wide range of tasks~\citep{NeurIPS'20:GPT3,NeurIPS'22:ChatGPT}. However, this progress comes at a price of growing computational cost: emerging techniques, including chain-of-thought~\citep{NeurIPS'22:CoT}, multi-step reasoning~\citep{Nature'25:DeepSeek-R1}, and long-horizon agentic tasks~\citep{SCIS'25:LLMAgentSurvey}, substantially increase the burden of inference.
To reduce the inference latency, \emph{speculative decoding}~\citep{ICML'23:Speculative,arxiv'23:speculative-sampling} has emerged as a widely adopted paradigm, in which a lightweight draft model rapidly generates a draft sequence that is then verified in parallel by a larger target model, achieving lossless acceleration~\citep{ICML'24:EAGLE,ICML'24:Lookahead,arxiv'25:Jacobi}.

Beyond standard autoregressive (AR) drafters, growing interest in the community has led several recent works~\citep{NAACL'25:SpecDiffusion,MLSys'26:SpecDiff2,ACL'26:DiffuSpec} to explore diffusion language models~(dLLMs)~\citep{NeurIPS'24:MaskedDiffusionLM,arxiv'25:LLaDA,arxiv'25:Dream} as non-autoregressive drafters, leveraging their \emph{bidirectional attention} and mask-and-denoise mechanism to generate multiple draft tokens simultaneously. SpecDiff~\citep{NAACL'25:SpecDiffusion} first demonstrated the feasibility of parallel diffusion drafting with few denoising steps; SpecDiff-2~\citep{MLSys'26:SpecDiff2} further introduced distillation mechanisms to align the diffusion drafter with the AR verifier. Building on top of these works, DFlash~\citep{ICML'26:DFlash} proposed a lightweight block diffusion drafter that generates all $k$ draft tokens through a single forward pass; crucially, it conditions each drafting layer on multi-layer hidden features from the target model, which improves the draft--target alignment and yields high acceptance rates.

In this work, we uncover a central pitfall that the \emph{global dependency} in diffusion drafters is a \emph{double-edged sword}. Specifically, global dependency arises from two components: bidirectional attention, which allows each draft token to attend to all other tokens in the draft block, and KV injection, which feeds the target hidden features into every drafting layer. While this global dependency enables global contextual modeling and one-pass parallel generation, it also introduces \emph{high variance} in draft quality. As illustrated in Figure~\ref{fig:observation}, this variance manifests at two levels: (i) \emph{domain-level} variance: the acceptance rate fluctuates significantly across task domains, much more so than that of AR drafters; and (ii) \emph{token-level} variance: the acceptance probability varies across draft positions, so the viable draft length changes drastically across rounds. Consequently, a static system with a fixed diffusion drafter and a constant candidate length suffers from degraded acceptance rates on out-of-distribution domains and wastes computation on tokens unlikely to be accepted.

To address these challenges, we propose \textsc{AdaFlash}, an online adaptive framework for diffusion-based speculative decoding with two components: (i)~an \emph{on-policy distillation}~(OPD) algorithm with reverse-KL and entry-wise divergence clipping, tailored specifically to diffusion drafters, which delivers stable convergence and continuously adapts the drafter to the deployment distribution; and (ii)~an \emph{adaptive length head} that dynamically adjusts the draft sequence length, substantially lowering the target model's verification cost with only a marginal decrease in acceptance length. We further introduce infrastructure improvements, including an asynchronous training--inference pipeline and adaptive request scheduling, that enable efficient online adaptation within a serving engine.

Experiments across eight benchmarks and three different foundation models, including both dense and mixture-of-experts architectures, validate the effectiveness of our \textsc{AdaFlash}. Both components yield significant performance gains, and our method consistently outperforms previous SOTA speculative decoding methods, including EAGLE-3, multi-token prediction, and DFlash, achieving up to a $5.3\times$ speedup over standard autoregressive decoding. The advantage is especially remarkable at high concurrency, where prior methods degrade below the standard AR baseline, while our \textsc{AdaFlash} sustains acceleration with up to 66\% higher average throughput than previous SOTA.

\section{Background and Key Observation}
\label{sec:problem}

This section introduces the background on speculative decoding and diffusion-based drafters, and then presents our key observation regarding the high variance issues of the diffusion drafters.

\subsection{Speculative Decoding and Diffusion Drafters}
\label{sec:preliminary}

\noindent\textbf{Speculative Decoding.~~}
Speculative decoding~\citep{ICML'23:Speculative,arxiv'23:speculative-sampling} accelerates inference of a large target model $p_{\v}$ by leveraging a smaller draft model $q_{\w}$. The draft model rapidly generates $k$ draft tokens, i.e., a \emph{draft sequence}, which is then verified in parallel by the target model.

Specifically, each draft token $x_i$ is accepted with probability $\min\{1, p_{\v}(x_i \mid \x_{<i}) / q_{\w}(x_i \mid \x_{<i})\}$; all tokens up to the first rejected one are retained, and a corrected token is resampled from the residual distribution at the rejection point. This sampling procedure guarantees that the output distribution is \emph{identical} to that of the target model, enabling lossless acceleration. Hereafter, acceptance length includes one target correction/bonus token. Following~\citet{ICML'23:Speculative}, its expectation per step is
$
    \E[\tau] = \frac{1 - \mathrm{Acc}^{k+1}}{1 - \mathrm{Acc}},
$
where $\mathrm{Acc} \triangleq \E_{x \sim q_{\w}}[\min\{1, {p_{\v}(x \mid \x_{<i})}/{q_{\w}(x \mid \x_{<i})}\}]$ is acceptance rate under the simplified i.i.d. assumption. The acceleration rate is then given by
\begin{equation*}
    \eta = \frac{\E[\tau]}{\rho k + 1},
\end{equation*}
where $k$ is the draft sequence's length (aka, candidate length), and $\rho \ll 1$ denotes the ratio of the draft model's inference time to that of the target model. This expression shows that the speedup is governed by the inference time ratio $\rho$ and the acceptance rate.

\vspace{1.5mm}
\noindent\textbf{Diffusion Language Models as Drafters.~~}
Existing speculative decoding methods predominantly employ AR drafters that \emph{sequentially} generate draft tokens, requiring $k$ sequential forward passes:
\begin{equation*}
    \textbf{for}\; i = 1, \ldots, k: \quad x_i \sim q_{\w}(\cdot \mid \x, x_1, \ldots, x_{i-1}).
\end{equation*}

Recently, several works~\citep{NAACL'25:SpecDiffusion,MLSys'26:SpecDiff2,ACL'26:DiffuSpec} have explored diffusion language models (dLLMs)~\citep{NeurIPS'24:MaskedDiffusionLM,arxiv'25:LLaDA,arxiv'25:Dream} as non-autoregressive drafters. Unlike AR models, dLLMs adopt a \emph{mask-and-denoise} strategy and leverage \emph{bidirectional attention}, enabling simultaneous prediction of multiple masked tokens without the sequential constraint of next-token prediction.

Building on this line, DFlash~\citep{ICML'26:DFlash} introduces a lightweight block diffusion drafter for speculative decoding. Specifically, given a prefix $\x$, the diffusion drafter initializes $k$ masked positions and generates \emph{all} draft tokens through a \emph{single} forward pass:
\begin{equation*}
    \{x_1, \ldots, x_k\} \sim q_{\w}(\cdot \mid \x, \z),
\end{equation*}
where $\z$ denotes the masked state of the draft block. Since the entire block is generated in one forward pass rather than $k$ sequential ones, the time ratio satisfies $\rho_{\mathrm{dLLM}} \approx \rho_{\mathrm{AR}} / k$, making the per-token drafting overhead roughly $k$ times smaller. This allows diffusion drafters to employ deeper, more expressive architectures without increasing latency, achieving both lower drafting cost and higher acceptance rates than AR drafters~\citep{ICML'26:DFlash}.

\begin{figure*}[t]
    \vspace{-1.5mm}
    \centering
    \begin{tabular}[b]{@{}c@{}}
        \includegraphics[height=3.7cm]{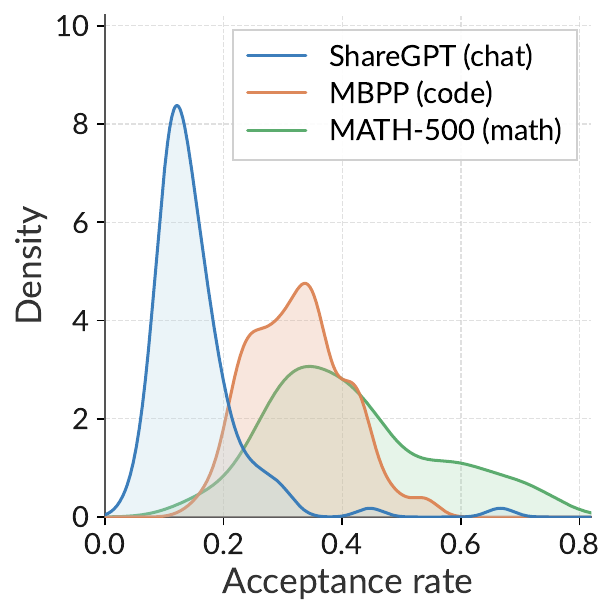} \\
        {~~~~~(a)}
    \end{tabular}
    \begin{tabular}[b]{@{}c@{}}
        \includegraphics[height=3.7cm]{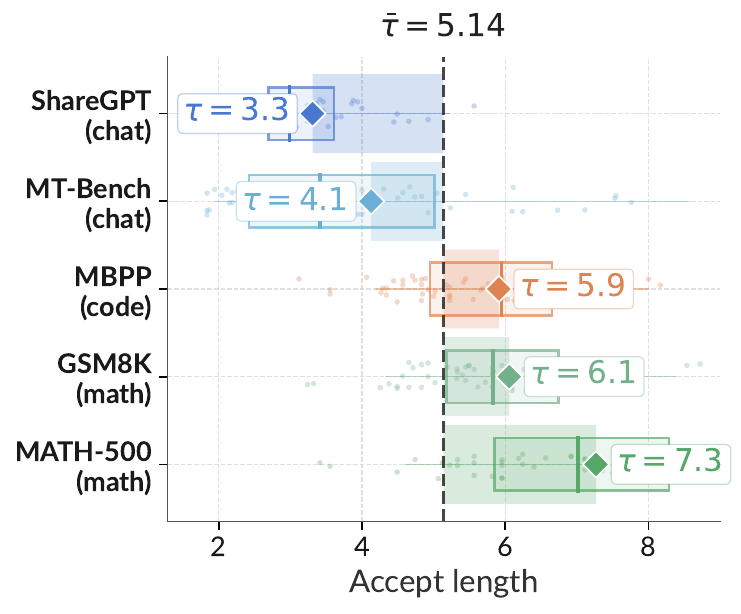} \\
        {~~~~~~~~~(b)}
    \end{tabular}\hspace{1mm}
    \begin{tabular}[b]{@{}c@{}}
        \includegraphics[height=3.7cm]{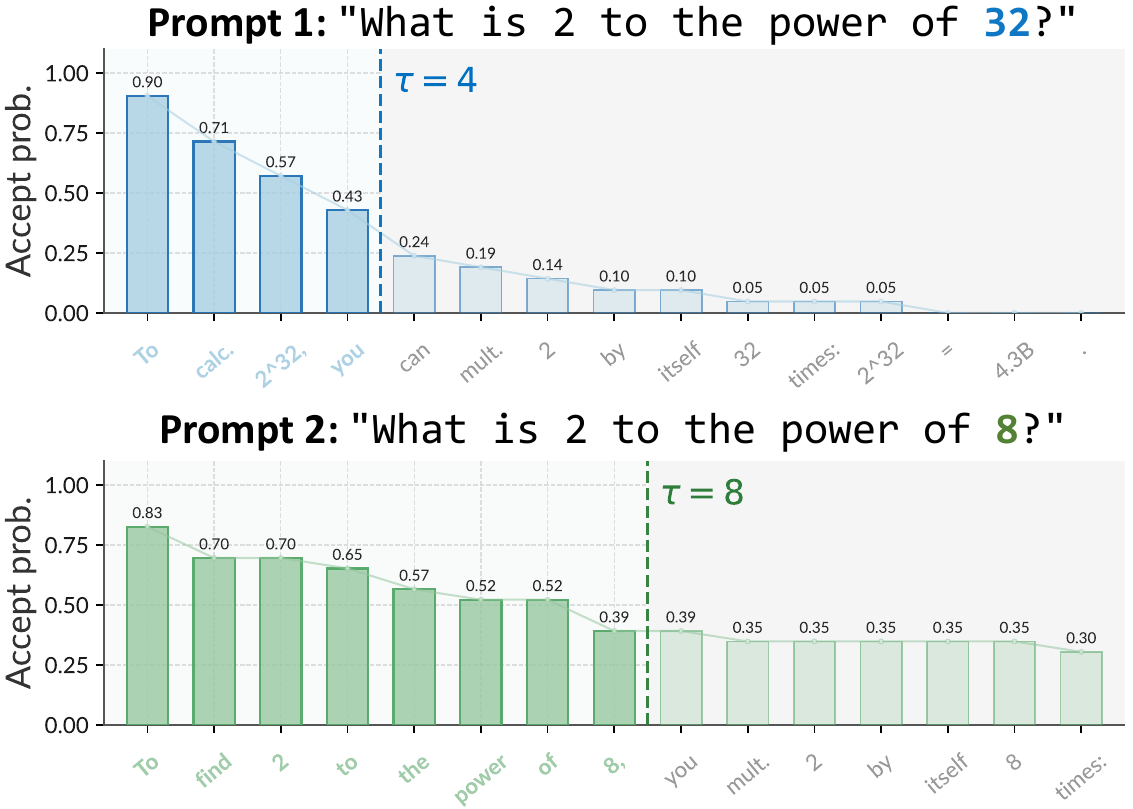} \\
        {~~~~~(c)}
    \end{tabular}
    \vspace{-2mm}
    \caption{Illustration of the \emph{high variance} issue in diffusion drafters. (a)~Probability density of the acceptance rate across three task domains (chat, code, and math), showing substantial distributional differences. (b)~Box plots of the acceptance length over five different datasets. (c)~Per-token acceptance probability for two nearly identical prefixes differing by only one token; the acceptance lengths diverge drastically, revealing high token-level variance of diffusion drafters.}
    \label{fig:observation}
    \vspace{-1mm}
\end{figure*}

\subsection{Key Observation: High Variance in Diffusion Drafters}
\label{sec:observation}

Despite the advantages of diffusion drafters described above, we uncover a central pitfall: the global dependency in diffusion drafters, arising from bidirectional attention and KV injection, is a \emph{double-edged sword}. Specifically, while it enables global contextual modeling and high-quality one-pass parallel generation, it also introduces \emph{high variance} in draft sequences (see Appendix~\ref{sec:additional_experiments:additional_results} for an ablation study). This variance manifests at the following two levels:
\vspace{-2mm}
\begin{itemize}[itemsep=1pt,leftmargin=1.5em]
    \item \emph{Domain-level variance}: The acceptance rate of the diffusion drafter fluctuates significantly across task domains (e.g., code, mathematics, dialogue), much more so than that of AR drafters. Because global dependency makes all draft tokens rely on each other and on the same target hidden features, the drafter becomes more sensitive to domain shifts at inference time. As shown in Figure~\ref{fig:observation}(a), the acceptance rate of DFlash varies substantially across domains.
    \item \emph{Token-level variance}: Within a single candidate sequence, the per-token divergence between the draft and target distributions exhibits considerable heterogeneity. Certain token positions closely match the target distribution, while others deviate substantially, leading to non-uniform acceptance patterns across positions. As shown in Figure~\ref{fig:observation}(c), the per-position acceptance probability fluctuates significantly.
\end{itemize}
\vspace{-1mm}

As illustrated in Figure~\ref{fig:observation}, both levels of variance are substantially large for diffusion drafters.
These two levels of variance pose distinct challenges for practical deployment. First, domain-level variance causes the acceptance rate to drop on out-of-distribution queries, degrading end-to-end speedup. A fixed, offline-trained diffusion drafter cannot adapt to the shifting query distribution encountered during deployment. Second, token-level variance leads to heterogeneous acceptance lengths across the scenarios of concurrent requests; if a fixed candidate length $k$ is used, requests with short acceptance lengths incur unnecessary verification cost, wasting target model computation on tokens unlikely to be accepted.

In summary, a static speculative system with a fixed diffusion drafter and a constant verification budget suffers efficiency degradation due to high variance issues. In the next section, we introduce our \textsc{AdaFlash} framework, which continuously updates the drafter parameters and adaptively adjusts the verification length to handle these two challenges, respectively.

\section{Our Approach}
\label{sec:approach}

In this section, we introduce our \textsc{AdaFlash} framework, which addresses the two challenges identified in Section~\ref{sec:problem}. We propose two components, on-policy distillation and an adaptive length head, to tackle the high variance in diffusion drafters.
\vspace{-2mm}
\begin{itemize}[itemsep=1pt,leftmargin=1.5em]
    \item \emph{On-Policy Distillation} continuously adapts the diffusion drafter to the target distribution during deployment, reducing the performance drop caused by domain-level variance.
    \item \emph{Adaptive Length Head} dynamically adjusts the verification length in each round based on the predicted acceptance rate, reducing the computational waste caused by token-level variance.
\end{itemize}
\vspace{-1mm}
Finally, we describe our \emph{infrastructure improvements} that realize these components using a serving engine, enabling asynchronous training and high-throughput serving.

\subsection{On-Policy Distillation for Diffusion Drafters}
\label{sec:opd-for-dllms}

In this part, we propose an on-policy distillation algorithm to address \emph{domain-level} variance.
As demonstrated earlier, the diffusion drafter exhibits a wide range of acceptance lengths across different domains.
Fortunately, during deployment, \emph{interactive feedback} is inherently available in speculative decoding: at each round, the target model produces its output distribution over the candidate positions, revealing exactly where the drafter diverges. Leveraging this feedback to update the drafter naturally forms a ``draft--feedback--adapt'' loop~\citep{ICML'24:OSD,ICML'26:OnlineSPEC}.

We exploit this feedback to perform \emph{on-policy distillation} (OPD): at each round $t$, the drafter $q_{\w_t}$ generates a candidate sequence of length $k$ given prefix $\x_{<t}$, and the target model returns its distributions $p_{\v}(\cdot \mid \x_{<t+i})$ at each position $i \in \{1, \ldots, k\}$ as supervision signals. 
Since the drafter is updated using online supervision derived from the target model's responses in the current deployment domain, the training distribution remains consistent with the inference distribution. This avoids the off-policy mismatch inherent in offline distillation and helps the drafter to adapt to the deployment distribution.
While this framework applies generally, the distinctive properties of diffusion drafters introduce two specific challenges, each of which we address with a dedicated mechanism detailed below.

\vspace{1.5mm}
\noindent\textbf{Mixture OPD Loss with Reverse-KL.~~}
Parallel drafting and bidirectional attention in diffusion drafters tend to produce high-entropy output distributions, i.e., the probability mass is spread broadly~\citep{ICML'22:huang22k}. However, speculative decoding only requires covering the high-probability regions of the target. Standard forward KL distillation (or cross-entropy loss) is mode-covering and forces the drafter to spread mass broadly, which is counterproductive in this setting. We therefore adopt \emph{reverse-KL} as the distillation objective, which penalizes the drafter for placing mass where the target has low probability, encouraging it to concentrate on the target's high-probability modes and directly improving acceptance rates.

To accelerate convergence toward the correct target model's prediction, we additionally include a hard-label cross-entropy loss on the target's top-1 token $x_i^\star = \arg\max_{x} p_{\v}(x \mid \x_{<t+i})$. The overall loss at round $t$ is
$
    \alpha \ell_{\mathrm{hard}} + (1-\alpha) \ell_{\mathrm{rkl}}
$,
where $\ell_{\mathrm{hard}} = -{1}/{k}\sum_{i=1}^{k} \log [q_{\w}(x_i^\star \mid \x_{<t}, \z)]_i$ is the hard-label cross-entropy, $\ell_{\mathrm{rkl}} = {1}/{k}\sum_{i=1}^{k} \KL \big([q_{\w}(\cdot \mid \x_{<t}, \z)]_i \,\|\, p_{\v}(\cdot \mid \x_{<t+i})\big)$ is the reverse-KL divergence, and $\alpha \in [0,1]$ is a mixing coefficient. The reverse-KL term drives the drafter's distribution toward the target's modes, while the hard-label term provides a low-variance gradient signal anchoring the drafter to the correct top-1 prediction.

\vspace{1.5mm}
\noindent\textbf{Entry-wise Divergence Clipping.~~}
The reverse-KL at each draft position $i$ is a sum $\sum_{y\in\mathcal{V}} q(y)\log(q(y)/p(y))$ over the entire vocabulary. A few entries $y$ in the vocabulary on which the drafter assigns non-negligible probability $q(y)$ but the target assigns a much smaller probability $p(y)$ yield a large ratio $q(y)/p(y)$; these entries can dominate the loss and produce large gradients that may mislead the overall update direction, as later illustrated in Figure~\ref{fig:div_clipping_example}.
To address this, we introduce an entry-wise divergence clipping mechanism that bounds each individual term, indexed by draft position $i$ and vocabulary entry $y$, at a threshold $\delta > 0$ before summation:
\begin{equation*}
    \ell_{\mathrm{rkl}}^{\mathrm{clip}} = \frac{1}{k}\sum_{i=1}^{k} \sum_{y \in \mathcal{V}} \min \bigg\{ [q_{\w}(y \mid \x_{<t}, \z)]_i \log \frac{[q_{\w}(y \mid \x_{<t}, \z)]_i}{p_{\v}(y \mid \x_{<t+i})}, \delta \bigg\},
\end{equation*}
where the $\min$ is applied entry-wise to each $(i, y)$ pair. This mitigates the outlier gradients produced by the few high-divergence vocabulary entries, while preserving the gradient signal from the remaining entries. The threshold $\delta$ controls the trade-off between robustness and convergence speed.
Combining both techniques, our final OPD loss is
\begin{equation*}
    \ell_{\mathrm{OPD}} = \alpha \, \ell_{\mathrm{hard}} + (1-\alpha) \, \ell_{\mathrm{rkl}}^{\mathrm{clip}}.
\end{equation*}
% At each round, the drafter parameters are updated via gradient descent on this loss. Since the feedback is obtained from the verification step that is already required by speculative decoding, OPD introduces negligible additional computational cost beyond the gradient update itself.

\subsection{Adaptive Length Head}
\label{sec:adaptive-length-head}

The previous section addresses domain-level variance by continuously adapting the drafter's parameters. We now tackle the complementary challenge of \emph{token-level variance}: existing methods use a fixed candidate length $k$, yet the per-position acceptance probability of diffusion drafters varies substantially, causing positions with low acceptance probability to waste the target model's computation. To this end, we introduce a lightweight \emph{adaptive length head} that predicts the acceptance ratio on the fly and truncates the candidate sequence to its viable prefix, eliminating wasteful verification of low-quality draft tokens.

\vspace{1.5mm}
\noindent \textbf{Design of the Adaptive Length Head.~~}
We attach a lightweight head to the diffusion drafter. At round $t$, the head receives the same input $\h_t \in \R^{k \times H}$ as the draft model. The head first applies a linear projection followed by SiLU activation, producing per-position features $\h'_i \in \R^{H'}$ for $i \in \{1, \ldots, k\}$, where $H' < H$. These features are then aggregated via mean pooling $\widehat{\h}_t = {1}/{k}\sum_{i=1}^k \h'_i$ to form a global representation that captures the overall draft quality. A final linear layer followed by sigmoid activation maps $\widehat{\h}_t$ to a scalar $\widehat{a}_t \in [0, 1]$, representing the predicted acceptance length of the draft sequence.
Given $\widehat{a}_t$, we determine the verification length $\widehat{k}_t$ as
\begin{equation*}
    \widehat{k}_t = \mathrm{clamp} \left(\lfloor \widehat{a}_t \cdot k \rfloor,\; 1,\; k\right).
\end{equation*}
After that, only the first $\widehat{k}_t$ tokens of the draft sequence are sent to the target model for verification. In practice, $\widehat{a}_t$ can be further scaled by a factor $\gamma \in \R^+$ to adapt to system load.

\vspace{1.5mm}
\noindent \textbf{Online Update of the Length Head.~~}
The length head is also continuously updated during deployment. Specifically, at each round $t$, after the target model verifies the candidate sequence, we obtain the ground-truth acceptance rate $a_t^\star \in [0, 1]$ (i.e., the fraction of accepted tokens). The length head is trained with a mean squared error (MSE) loss:
\begin{equation*}
    \ell_{\mathrm{len}} = \left(\widehat{a}_t - a_t^\star\right)^2.
\end{equation*}
Since the verification outcome is already available from the speculative decoding pipeline, this supervision signal incurs no additional cost.
We remark that the length head's loss is \emph{detached} from the drafter, so that $\ell_{\mathrm{len}}$ updates only the length head parameters and no gradient flows back into the drafter. This ensures that the MSE objective of the length head and the reverse-KL objective of OPD remain fully decoupled, preventing interference during training. During deployment progresses, the length head adapts to the evolving diffusion drafter and the changing distribution, continuously refining its predictions.

% The length head parameters are updated asynchronously with respect to the main drafter, ensuring that the overhead is minimal and does not block the generation pipeline. As deployment progresses, the length head adapts to the evolving drafter (improved by OPD) and the changing query distribution, continuously refining its predictions.

\begin{myRemark}[Comparison with DSpark]
    A concurrent work, DSpark~\citep{misc'26:DSpark}, also dynamically adjusts the verification length via a per-position confidence head coupled with a hardware-aware scheduler. Our approach differs in two aspects: (i) DSpark estimates per-position conditional acceptance probabilities and requires post-hoc calibration to correct for compounding errors, whereas our adaptive length head directly predicts the overall acceptance length, offering a simpler formulation that bypasses draft sequence's probability calibration; (ii) DSpark's confidence head is trained offline and remains fixed during inference, while our adaptive length head continuously evolves on the fly, co-adapting with the evolving diffusion drafter during deployment, with an empirical comparison in Appendix~\ref{sec:additional_experiments:additional_results}. Beyond these differences, DSpark's hardware-aware scheduling is complementary to our approach and can be integrated into our \textsc{AdaFlash} framework for further improvement.
\end{myRemark}

\subsection{Infrastructure Design for Online Adaptation}
\label{sec:system-design}

In previous sections, we have introduced OPD with divergence clipping and the adaptive length head to address the high variance of diffusion drafters. We now describe the infrastructure design that realizes these components efficiently via a serving engine built on SGLang~\citep{NeurIPS'24:sglang}. Two requirements must be met: (i)~the training process must run asynchronously without blocking inference, and (ii)~variable verification lengths must be scheduled efficiently under dynamic GPU memory constraints. We address both with the following system design.

\vspace{1.5mm}
\noindent \textbf{Asynchronous Training--Inference Pipeline.~~}
We decouple the system into an \emph{inference server} and a \emph{training server} connected through a shared data buffer. During inference, the speculative decoding pipeline records the on-policy generation trajectories into the buffer, comprising the input prompts and the drafter's responses. The training server asynchronously leverages these trajectories and computes the target model's supervision signals, and updates the drafter and length head parameters using our OPD loss. This decoupling ensures that training never blocks token generation.

Once the training server produces updated weights, the inference server incorporates them via \emph{lightweight hot-reloading}: only the drafter and length head parameters are swapped into GPU memory, while the target model remains untouched. This operation completes between scheduling steps, introducing negligible latency.

\vspace{1.5mm}
\noindent \textbf{Adaptive Request Scheduling.~~}
Existing serving frameworks (e.g., SGLang~\citep{NeurIPS'24:sglang}) typically use a fixed batch shape $\text{bs} \times k$ for verification, where all requests share the same candidate length $k$. To support our adaptive verification lengths, we modify the serving engine to accept variable candidate lengths and pack them into a compact verification batch whose total length equals $\sum_i^N \widehat{k}_i$, thereby reducing computational waste and improving GPU utilization under high concurrency.

Since the per-request verification length varies, the number of requests that can be processed per batch becomes less predictable. To address this, we apply an \emph{exponential moving average} (EMA) to dynamically estimate the expected number of requests to process. Concretely, after each scheduling round, we update the estimate as
$
    \widehat{N} \leftarrow (1 - \beta)\,\widehat{N} + \beta\,N,
$
where $N$ is the number of requests actually processed in the current round and $\beta \in (0,1)$ is the smoothing coefficient. The scheduler uses $\widehat{N}$ to determine how many requests to admit in subsequent rounds. When the actual memory usage still exceeds the budget, the scheduler returns excess requests to the pending queue. This strategy maximizes GPU utilization while preventing out-of-memory failures, and naturally accommodates our \textsc{AdaFlash} framework where the drafter evolves during deployment.

\section{Experiments}
\label{sec:experiment}

To comprehensively evaluate the effectiveness of \textsc{AdaFlash}, we conduct experiments across eight datasets including two long-sequence reasoning benchmarks, and three target models, under various concurrency levels. Our experimental evaluation aims to answer the following research questions:

\vspace{-2mm}
\begin{itemize}[itemsep=2pt,leftmargin=2.2em,labelwidth=*,align=left]
    \item [\textbf{Q1:}] Does the diffusion drafter suffer from high variance during deployment, and does this affect the overall speedup under different concurrency levels?
    \item [\textbf{Q2:}] Can \textsc{AdaFlash} improve end-to-end inference throughput by leveraging on-policy distillation and adaptive verification length?
    \item [\textbf{Q3:}] Is each component of \textsc{AdaFlash} effective? How do the hyperparameter choices (e.g., clipping threshold, mixing coefficient) affect the performance?
\end{itemize}
% \vspace{-2mm}

\subsection{Experimental Setup}
\label{sec:experimental_setup}
In this part, we introduce the experimental setup, including the datasets and contenders.

\vspace{1mm}
\noindent \textbf{Datasets.~~} We conduct experiments on eight benchmark datasets spanning diverse domains: two math reasoning tasks, \emph{MathQA}~\citep{NAACL'19:MathQA} and \emph{GSM8K}~\citep{arxiv'21:GSM8K}; two code generation tasks, \emph{OpenCodeInstruct}~\citep{arxiv'25:OpenCodeInstruct} and \emph{CodeAlpaca}~\citep{codealpaca}; a general-purpose dialogue dataset \emph{ShareGPT}; and a mixed-domain dataset \emph{Blend} that combines samples from the above domains to evaluate cross-domain robustness. We also evaluate on two long-sequence reasoning benchmarks, \emph{MATH-500}~\citep{arxiv'23:LetsVerifyStep} and \emph{AIME25}~\citep{misc'25:aime25}, where the maximum output length is set to $32{,}768$ tokens with thinking mode enabled, to assess performance on extended generation tasks.

\vspace{1mm}
\noindent \textbf{Contenders.~~} We compare \textsc{AdaFlash} with the following methods: \rom{1} \emph{Standard AR} decoding (autoregressive generation without speculative decoding). \rom{2} \emph{EAGLE-3}~\citep{arxiv'25:EAGLE-3}, a state-of-the-art autoregressive drafter that directly predicts tokens from fused multi-layer target-model features and employs tree-structured verification. \rom{3} \emph{DFlash}~\citep{ICML'26:DFlash}, the offline diffusion drafter that generates candidate sequences through parallel denoising. \rom{4} \emph{OSD}~\citep{ICML'24:OSD}, which applies online distillation to continuously update the drafter. We apply the OSD forward-KL online-update rule to the DFlash drafter. \textsc{AdaFlash} extends DFlash with on-policy distillation for diffusion drafters and an adaptive length head. Due to page limit, we defer the implementation details to Appendix~\ref{sec:implementation_details}.

\begin{table*}[t]
    \vspace{-1mm}
    \centering
    \caption{Performance comparison of different speculative decoding methods on six benchmark datasets using Q3-8B (short for Qwen3-8B) and Q3-30B (short for Qwen3-Coder-30B-A3B) as target models. We report \textsc{Speedup}  (relative to standard AR decoding) and average accepted length $\tau$  under four concurrency levels ($C \in \{1, 32, 64, 128\}$). The best results are highlighted in \textbf{bold}.}
    \vspace{-2mm}
    \label{tab:performance_comparison}
    \resizebox{\textwidth}{!}{
        \begin{tabular}{lcc *{7}{cc}}
            \toprule
            \multirow{3}{*}{Model} & \multirow{3}{*}{Method}            & \multirow{3}{*}{Conc.} & \multicolumn{4}{c}{\textsc{Math}} & \multicolumn{4}{c}{\textsc{Code}} & \multicolumn{2}{c}{\textsc{Chat}}    & \multicolumn{2}{c}{\textsc{Hybrid}} & \multicolumn{2}{c}{}                                                                                                                                                                                     \\
            \cmidrule(lr){4-7} \cmidrule(lr){8-11} \cmidrule(lr){12-13} \cmidrule(lr){14-15}
                                   &                                    &                        & \multicolumn{2}{c}{MathQA}        & \multicolumn{2}{c}{GSM8K}         & \multicolumn{2}{c}{OpenCodeInstruct} & \multicolumn{2}{c}{CodeAlpaca}      & \multicolumn{2}{c}{ShareGPT} & \multicolumn{2}{c}{Blend} & \multicolumn{2}{c}{\textit{Avg.}}                                                                                                             \\
            \cmidrule(lr){4-17}
                                   &                                    &                        & Speedup                           & $\tau$                            & Speedup                              & $\tau$                              & Speedup                      & $\tau$                    & Speedup                           & $\tau$ & Speedup               & $\tau$ & Speedup               & $\tau$ & Speedup               & $\tau$ \\
            \midrule
            \multirow{16}{*}{Q3-8B}
                                   & \multirow{4}{*}{EAGLE-3}           & 1                      & 2.45$\times$                      & 4.60                              & 2.55$\times$                         & 4.75                                & 2.20$\times$                 & 4.06                      & 2.37$\times$                      & 4.58   & 2.09$\times$          & 3.89   & 2.36$\times$          & 4.54   & 2.34$\times$          & 4.40   \\
                                   &                                    & 32                     & 0.72$\times$                      & 4.60                              & 0.70$\times$                         & 4.76                                & 0.68$\times$                 & 4.07                      & 0.67$\times$                      & 4.58   & 0.61$\times$          & 3.90   & 0.70$\times$          & 4.54   & 0.68$\times$          & 4.41   \\
                                   &                                    & 64                     & 0.46$\times$                      & 4.59                              & 0.43$\times$                         & 4.75                                & 0.44$\times$                 & 4.07                      & 0.40$\times$                      & 4.57   & 0.40$\times$          & 3.90   & 0.45$\times$          & 4.54   & 0.43$\times$          & 4.40   \\
                                   &                                    & 128                    & 0.34$\times$                      & 4.59                              & 0.35$\times$                         & 4.75                                & 0.33$\times$                 & 4.07                      & 0.32$\times$                      & 4.58   & 0.30$\times$          & 3.90   & 0.33$\times$          & 4.54   & 0.33$\times$          & 4.41   \\
            \cmidrule(l){2-17}
                                   & \multirow{4}{*}{DFlash}            & 1                      & 4.38$\times$                      & 7.09                              & 3.84$\times$                         & 6.27                                & 4.08$\times$                 & 6.69                      & 3.39$\times$                      & 5.63   & 2.11$\times$          & 3.40   & 3.39$\times$          & 6.07   & 3.53$\times$          & 5.86   \\
                                   &                                    & 32                     & 1.86$\times$                      & 7.10                              & 1.59$\times$                         & 6.28                                & 1.77$\times$                 & 6.68                      & 1.46$\times$                      & 5.66   & 1.05$\times$          & 3.42   & 1.51$\times$          & 6.04   & 1.54$\times$          & 5.86   \\
                                   &                                    & 64                     & 1.24$\times$                      & 7.11                              & 1.03$\times$                         & 6.29                                & 1.17$\times$                 & 6.67                      & 0.91$\times$                      & 5.65   & 0.70$\times$          & 3.41   & 0.98$\times$          & 6.06   & 1.01$\times$          & 5.86   \\
                                   &                                    & 128                    & 0.90$\times$                      & 7.11                              & 0.82$\times$                         & 6.28                                & 0.86$\times$                 & 6.67                      & 0.74$\times$                      & 5.66   & 0.51$\times$          & 3.40   & 0.74$\times$          & 6.04   & 0.76$\times$          & 5.86   \\
            \cmidrule(l){2-17}
                                   & \multirow{4}{*}{OSD}               & 1                      & 5.13$\times$                      & 9.44                              & 4.33$\times$                         & 7.45                                & 4.78$\times$                 & 8.21                      & 3.65$\times$                      & 6.59   & 2.18$\times$          & 3.78   & 3.60$\times$          & 6.83   & 3.95$\times$          & 7.05   \\
                                   &                                    & 32                     & 2.16$\times$                      & 9.42                              & 1.80$\times$                         & 7.47                                & 2.03$\times$                 & 8.20                      & 1.55$\times$                      & 6.61   & 1.08$\times$          & 3.79   & 1.57$\times$          & 6.82   & 1.70$\times$          & 7.05   \\
                                   &                                    & 64                     & 1.45$\times$                      & 9.44                              & 1.16$\times$                         & 7.43                                & 1.34$\times$                 & 8.20                      & 0.97$\times$                      & 6.60   & 0.72$\times$          & 3.80   & 1.07$\times$          & 6.85   & 1.12$\times$          & 7.05   \\
                                   &                                    & 128                    & 1.05$\times$                      & 9.47                              & 0.92$\times$                         & 7.44                                & 0.97$\times$                 & 8.20                      & 0.78$\times$                      & 6.62   & 0.53$\times$          & 3.81   & 0.76$\times$          & 6.82   & 0.83$\times$          & 7.06   \\
            \cmidrule(l){2-17}
                                   & \multirow{4}{*}{\textsc{AdaFlash}} & 1                      & \textbf{5.32}$\times$             & 9.83                              & \textbf{4.54}$\times$                & 7.75                                & \textbf{4.87}$\times$        & 8.40                      & \textbf{3.75}$\times$             & 6.80   & \textbf{2.24}$\times$ & 3.92   & \textbf{3.66}$\times$ & 7.00   & \textbf{4.06}$\times$ & 7.28   \\
                                   &                                    & 32                     & \textbf{2.27}$\times$             & 9.84                              & \textbf{1.83}$\times$                & 7.76                                & \textbf{2.03}$\times$        & 8.39                      & \textbf{1.58}$\times$             & 6.79   & \textbf{1.10}$\times$ & 3.92   & \textbf{1.61}$\times$ & 6.97   & \textbf{1.74}$\times$ & 7.28   \\
                                   &                                    & 64                     & \textbf{1.65}$\times$             & 9.51                              & \textbf{1.32}$\times$                & 7.28                                & \textbf{1.51}$\times$        & 8.02                      & \textbf{1.18}$\times$             & 6.31   & \textbf{1.04}$\times$ & 3.61   & \textbf{1.26}$\times$ & 6.56   & \textbf{1.33}$\times$ & 6.88   \\
                                   &                                    & 128                    & \textbf{1.34}$\times$             & 9.51                              & \textbf{1.27}$\times$                & 7.27                                & \textbf{1.25}$\times$        & 8.05                      & \textbf{1.11}$\times$             & 6.30   & \textbf{0.87}$\times$ & 3.58   & \textbf{1.05}$\times$ & 6.55   & \textbf{1.15}$\times$ & 6.88   \\
            \midrule
            \multirow{16}{*}{Q3-30B}
                                   & \multirow{4}{*}{EAGLE-3}           & 1                      & 1.79$\times$                      & 4.39                              & 1.56$\times$                         & 4.06                                & 1.96$\times$                 & 5.15                      & 1.93$\times$                      & 4.94   & 1.32$\times$ & 3.08   & 1.79$\times$          & 4.49   & 1.73$\times$          & 4.35   \\
                                   &                                    & 32                     & 1.33$\times$                      & 4.40                              & 1.15$\times$                         & 4.06                                & 2.85$\times$                 & 5.15                      & 1.71$\times$                      & 4.93   & 1.15$\times$          & 3.08   & 1.49$\times$          & 4.50   & 1.61$\times$          & 4.35   \\
                                   &                                    & 64                     & 1.03$\times$                      & 4.40                              & 0.86$\times$                         & 4.06                                & 1.44$\times$                 & 5.15                      & 1.33$\times$                      & 4.93   & 0.91$\times$          & 3.08   & 1.16$\times$          & 4.48   & 1.12$\times$          & 4.35   \\
                                   &                                    & 128                    & 0.71$\times$                      & 4.41                              & 0.63$\times$                         & 4.06                                & 1.15$\times$                 & 5.15                      & 1.10$\times$                      & 4.93   & 0.60$\times$          & 3.08   & 0.77$\times$          & 4.49   & 0.83$\times$          & 4.35   \\
            \cmidrule(l){2-17}
                                   & \multirow{4}{*}{DFlash}            & 1                      & 2.09$\times$                      & 5.15                              & 1.84$\times$                         & 4.85                                & 2.56$\times$                 & 6.86                      & 2.39$\times$                      & 6.03   & 1.04$\times$          & 2.79   & 2.07$\times$          & 5.43   & 2.00$\times$          & 5.19   \\
                                   &                                    & 32                     & 1.61$\times$                      & 5.14                              & 1.41$\times$                         & 4.85                                & 3.80$\times$                 & 6.84                      & 2.15$\times$                      & 6.01   & 1.14$\times$          & 2.79   & 1.74$\times$          & 5.42   & 1.98$\times$          & 5.18   \\
                                   &                                    & 64                     & 1.26$\times$                      & 5.16                              & 1.05$\times$                         & 4.84                                & 1.95$\times$                 & 6.84                      & 1.68$\times$                      & 6.01   & 0.93$\times$          & 2.81   & 1.39$\times$          & 5.42   & 1.38$\times$          & 5.18   \\
                                   &                                    & 128                    & 0.87$\times$                      & 5.15                              & 0.77$\times$                         & 4.84                                & 1.55$\times$                 & 6.84                      & 1.40$\times$                      & 6.00   & 0.60$\times$          & 2.80   & 0.92$\times$          & 5.42   & 1.02$\times$          & 5.18   \\
            \cmidrule(l){2-17}
                                   & \multirow{4}{*}{OSD}               & 1                      & 2.79$\times$                      & 8.38                              & 2.25$\times$                         & 6.44                                & 2.96$\times$                 & 8.28                      & 2.58$\times$                      & 7.09   & 1.30$\times$          & 3.48   & 2.24$\times$          & 6.38   & 2.35$\times$          & 6.68   \\
                                   &                                    & 32                     & 2.04$\times$                      & 8.39                              & 1.60$\times$                         & 6.43                                & 4.30$\times$                 & 8.27                      & 2.26$\times$                      & 6.94   & 1.23$\times$          & 3.35   & 1.83$\times$          & 6.37   & 2.21$\times$          & 6.63   \\
                                   &                                    & 64                     & 1.59$\times$                      & 8.36                              & 1.26$\times$                         & 6.43                                & 2.21$\times$                 & 8.27                      & 1.77$\times$                      & 6.95   & 0.98$\times$          & 3.37   & 1.43$\times$          & 6.35   & 1.54$\times$          & 6.62   \\
                                   &                                    & 128                    & 1.11$\times$                      & 8.39                              & 0.89$\times$                         & 6.43                                & 1.75$\times$                 & 8.27                      & 1.47$\times$                      & 6.95   & 0.65$\times$          & 3.39   & 0.99$\times$          & 6.35   & 1.14$\times$          & 6.63   \\
            \cmidrule(l){2-17}
                                   & \multirow{4}{*}{\textsc{AdaFlash}} & 1                      & \textbf{2.86}$\times$             & 8.53                              & \textbf{2.32}$\times$                & 6.69                                & \textbf{2.99}$\times$        & 8.46                      & \textbf{2.65}$\times$             & 7.21   & \textbf{1.32}$\times$ & 3.52   & \textbf{2.27}$\times$ & 6.56   & \textbf{2.40}$\times$ & 6.83   \\
                                   &                                    & 32                     & \textbf{2.05}$\times$             & 8.53                              & \textbf{1.59}$\times$                & 6.69                                & \textbf{4.32}$\times$        & 8.43                      & \textbf{2.33}$\times$             & 7.08   & \textbf{1.23}$\times$ & 3.42   & \textbf{1.85}$\times$ & 6.53   & \textbf{2.23}$\times$ & 6.78   \\
                                   &                                    & 64                     & \textbf{1.60}$\times$             & 8.08                              & \textbf{1.32}$\times$                & 6.42                                & \textbf{2.35}$\times$        & 7.96                      & \textbf{1.98}$\times$             & 6.63   & \textbf{1.08}$\times$ & 3.12   & \textbf{1.51}$\times$ & 6.16   & \textbf{1.64}$\times$ & 6.40   \\
                                   &                                    & 128                    & \textbf{1.57}$\times$             & 8.11                              & \textbf{1.28}$\times$                & 6.43                                & \textbf{2.57}$\times$        & 7.96                      & \textbf{2.32}$\times$             & 6.59   & \textbf{1.04}$\times$ & 3.11   & \textbf{1.33}$\times$ & 6.14   & \textbf{1.69}$\times$ & 6.39   \\
            \bottomrule
        \end{tabular}
    }
    \vspace{-1.8mm}
\end{table*}

\subsection{Evaluation of Our Approach}
\label{sec:exp:results}

In this part, to answer \textbf{Q1} and \textbf{Q2}, we evaluate our \textsc{AdaFlash} framework across a range of setups, including different datasets, target models, and concurrency configurations.

\vspace{1mm}
\noindent \textbf{Validating the High Variance of Diffusion Drafters.~~}
To answer \textbf{Q1}, we first examine whether diffusion drafters exhibit higher variance than autoregressive drafters during deployment, as hypothesized in Section~\ref{sec:problem}. As shown in Table~\ref{tab:performance_comparison}, compared with EAGLE-3 in our evaluated benchmarks, diffusion drafter exhibits a wider range of acceptance lengths across domains: for Qwen3-8B, $\tau$ ranges from 3.40 on ShareGPT to 7.09 on MathQA, a ratio of approximately $2.1\times$. In contrast, EAGLE-3 exhibits a much narrower range, from 3.89 to 4.75 (ratio $\approx 1.2\times$). This confirms that diffusion drafters exhibit larger cross-domain variation in draft quality, justifying the domain-level variance introduced in Section~\ref{sec:observation}. Moreover, this large variance in acceptance length directly translates into wasted verification cost: with a fixed candidate length $k=15$, the target model verifies all tokens even when only a few are likely to be accepted, degrading throughput under high concurrency. These observations motivate the two components of \textsc{AdaFlash}: on-policy distillation to adapt the drafter to each deployment domain, and the adaptive length head to reduce verification waste.

\vspace{1mm}
\noindent \textbf{Evaluation Results on Different Setups.~~}
To answer \textbf{Q2}, Table~\ref{tab:performance_comparison} summarizes the speedup and average acceptance length across six datasets, two target models, and four concurrency levels ($C \in \{1, 32, 64, 128\}$). Across all settings, \textsc{AdaFlash} consistently achieves the highest speedup. For Qwen3-8B at $C=1$, \textsc{AdaFlash} attains an average speedup of $4.06\times$, compared to $3.53\times$ for DFlash, $3.95\times$ for OSD, and $2.34\times$ for EAGLE-3. The acceptance length also improves consistently: on MathQA, $\tau$ increases from $7.09$ (DFlash) to $9.44$ (OSD) to $9.83$ (\textsc{AdaFlash}), confirming that our on-policy distillation is more effective than the standard knowledge distillation used by OSD. These improvements generalize across model architectures and scales: on the mixture-of-experts model Qwen3-Coder-30B-A3B, \textsc{AdaFlash} also consistently achieves the highest speedup.

\begin{figure*}[t]
    % \vspace{-1mm}
    \centering
    \begin{tabular}[b]{@{}c@{}}
        \includegraphics[height=3.4cm]{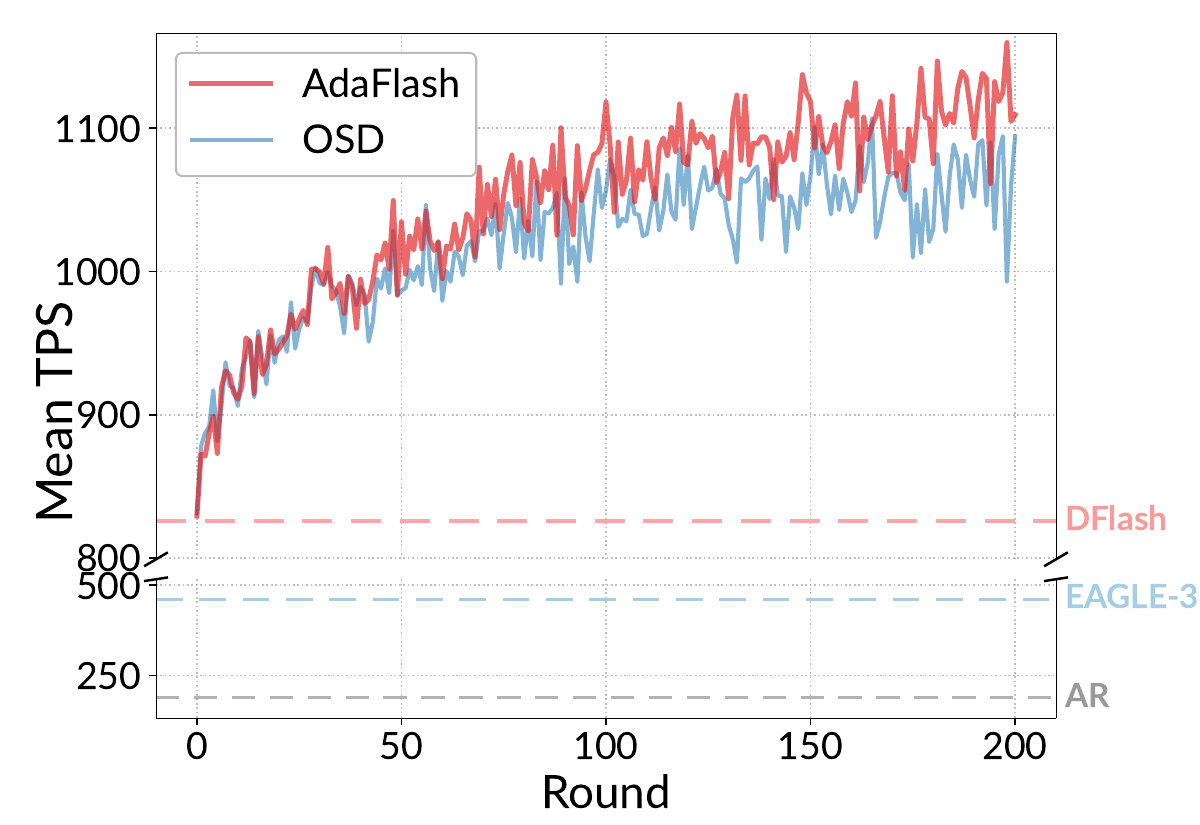} \\
        {~~~~~(a)}
    \end{tabular}
    \begin{tabular}[b]{@{}c@{}}
        \includegraphics[height=3.4cm]{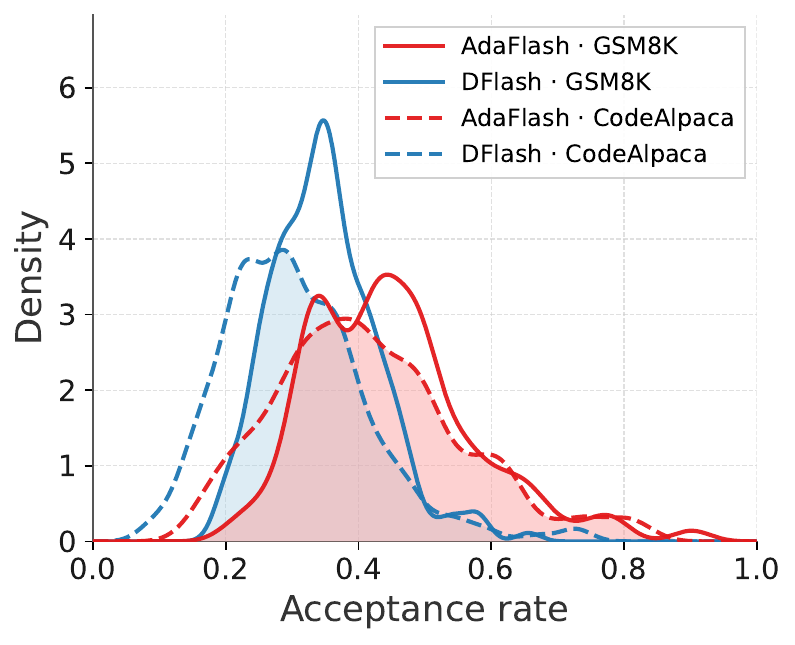} \\
        {~~~~~~~~~(b)}
    \end{tabular}
    \begin{tabular}[b]{@{}c@{}}
        \includegraphics[height=3.4cm]{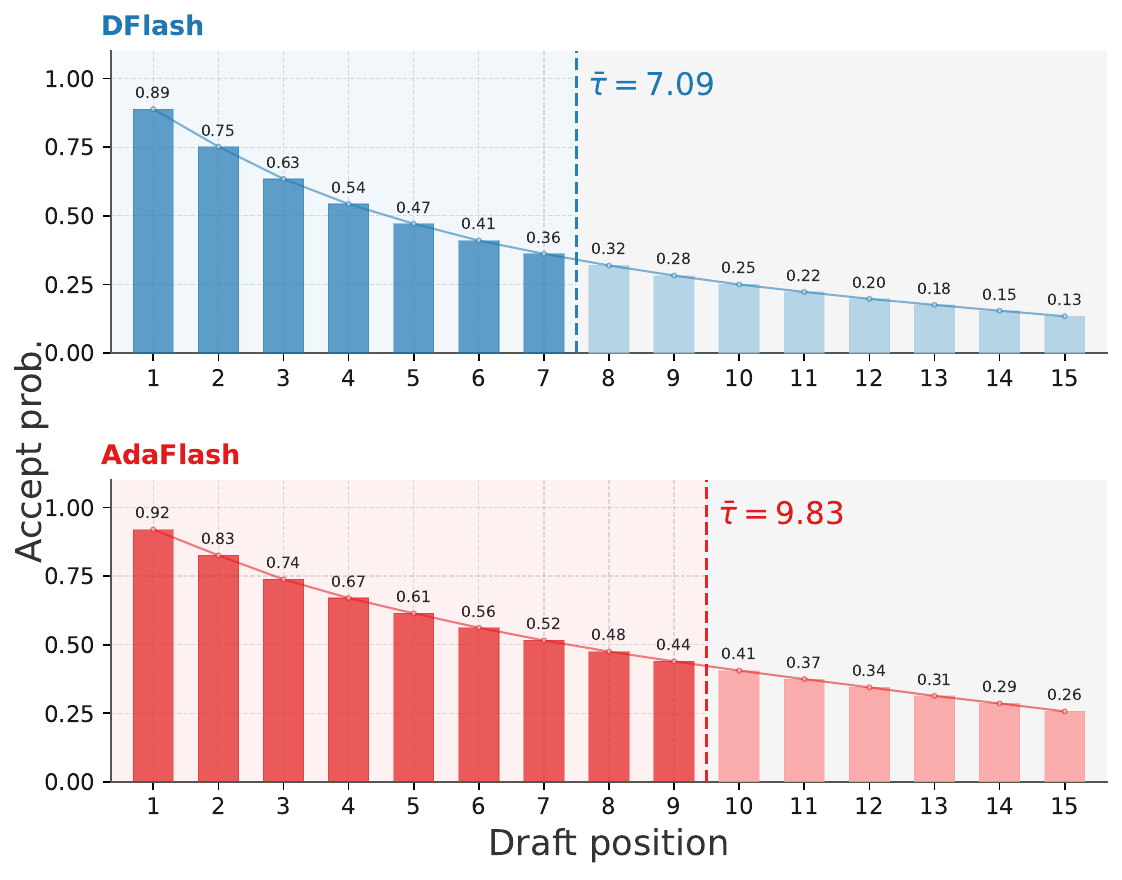} \\
        {~~~~~(c)}
    \end{tabular}
    \vspace{-2mm}
    \caption{Analysis of \textsc{AdaFlash} using Qwen3-8B as target model: (a)~Mean throughput (TPS) vs.\ online adaptation rounds on the MathQA dataset, showing continuous improvement over time. (b)~Per-sample acceptance rate distributions on GSM8K and CodeAlpaca, both shifted toward higher rates by on-policy distillation. (c)~Per-position acceptance probability on the MathQA dataset.}
    \label{fig:exp_analysis}
    % \vspace{-2mm}
\end{figure*}

The advantage of \textsc{AdaFlash} becomes more remarkable under high concurrency. At $C=128$ with Qwen3-8B, \textsc{AdaFlash} maintains a speedup of $1.15\times$, while OSD and DFlash fall below standard decoding, indicating that methods with a fixed verification length waste GPU compute when only a few draft tokens are accepted. In contrast, the adaptive length head reduces the verification length and thus the per-round verification cost. On Qwen3-Coder-30B-A3B at $C=128$, \textsc{AdaFlash} achieves an average speedup of $1.69\times$ compared to $1.02\times$ for DFlash, corresponding to 66\% higher throughput. Furthermore, Figure~\ref{fig:exp_analysis}(a) shows that throughput improves progressively over online adaptation rounds, with \textsc{AdaFlash} consistently outperforming OSD throughout deployment.

\vspace{1mm}
\noindent \textbf{Effectiveness of \textsc{AdaFlash}.~~}
We next examine how \textsc{AdaFlash} mitigates the two types of variance. For domain-level variance, Figure~\ref{fig:exp_analysis}(b) plots per-sample acceptance rate distributions on GSM8K and CodeAlpaca: DFlash concentrates at low rates with little cross-domain overlap, whereas \textsc{AdaFlash} shifts both distributions toward higher rates with substantially greater overlap, confirming that on-policy distillation raises the overall acceptance rate while narrowing the domain gap. For token-level variance, Figure~\ref{fig:exp_analysis}(c) shows that the per-position acceptance probability of DFlash decreases monotonically along the draft sequence, with later positions exhibiting substantially lower rates. On-policy distillation lifts the acceptance probability at every position, particularly at later positions where the baseline drafter suffers the most, thereby narrowing the gap between early and late positions and extending the average acceptance length from $7.09$ to $9.83$.

\subsection{Additional Experimental Results}
\label{sec:exp:other_analysis}

\noindent \textbf{Ablation Study.~~}
To answer \textbf{Q3}, we conduct ablation studies to validate each component's contribution in \textsc{AdaFlash}. As shown in Table~\ref{tab:ablation}, we progressively add components and evaluate on \textsc{GSM8K} with Qwen3-8B as the target model.

\begin{table*}[t]
    \centering
    \vspace{-1mm}
    \caption{Ablation study of \textsc{AdaFlash} components on the \textsc{GSM8K} dataset using Qwen3-8B as the target model. We evaluate four components: divergence clipping, mixture OPD loss, adaptive length head, and online update, and report the average accepted length $\tau$ and \textsc{Speedup} at concurrency levels $C \in \{1, 128\}$. The best results are highlighted in \textbf{bold}.}
    \vspace{-2mm}
    \label{tab:ablation}
    \resizebox{0.99\textwidth}{!}{%
        \begin{small}
            \setlength{\tabcolsep}{14pt}
            \begin{tabular}{cc  cc  ccc}
                \toprule
                \multicolumn{2}{c}{\textbf{On-Policy Distillation}} & \multicolumn{2}{c}{\textbf{Adaptive Length Head}} & \multicolumn{3}{c}{\textbf{GSM8K}}                                                                     \\
                \cmidrule(lr){1-2} \cmidrule(lr){3-4} \cmidrule(lr){5-7}
                \makecell{\textbf{Divergence}                                                                                                                                                                                    \\\textbf{Clipping}} & \makecell{\textbf{Mixture}\\\textbf{OPD Loss}} & \makecell{\textbf{Length}\\\textbf{Head}} & \makecell{\textbf{Online}\\\textbf{Update}} & \textbf{Conc.} & $\boldsymbol{\tau}$ & \textbf{Speedup} \\
                \midrule
                \multirow{2}{*}{}                                   & \multirow{2}{*}{}                                 & \multirow{2}{*}{\checkmark}        & \multirow{2}{*}{\checkmark}
                                                                    & 1                                                 & 7.389                              & 4.17$\times$                                                      \\
                                                                    &                                                   &                                    &                             & 128 & 6.802 & 1.18$\times$          \\
                \multirow{2}{*}{\checkmark}                         & \multirow{2}{*}{}                                 & \multirow{2}{*}{\checkmark}        & \multirow{2}{*}{\checkmark}
                                                                    & 1                                                 & 7.590                              & 4.31$\times$                                                      \\
                                                                    &                                                   &                                    &                             & 128 & 7.212 & 1.25$\times$          \\
                \multirow{2}{*}{\checkmark}                         & \multirow{2}{*}{\checkmark}                       & \multirow{2}{*}{\checkmark}        & \multirow{2}{*}{\checkmark}
                                                                    & 1                                                 & 7.751                              & \textbf{4.54$\times$}                                             \\
                                                                    &                                                   &                                    &                             & 128 & 7.270 & \textbf{1.27$\times$} \\
                \cmidrule(lr){1-7}
                \checkmark                                          & \checkmark                                        &                                    &                             & 128 & 6.594 & 1.00$\times$          \\
                \checkmark                                          & \checkmark                                        & \checkmark                         &                             & 128 & 7.652 & 1.21$\times$          \\
                \checkmark                                          & \checkmark                                        & \checkmark                         & \checkmark                  & 128 & 7.270 & \textbf{1.27$\times$} \\
                \bottomrule
            \end{tabular}
        \end{small}
    }
    % \vspace{-1mm}
\end{table*}

\begin{table*}[t]
    \vspace{-3.5mm}
    \centering
    \newcommand{\hyperparamtab}[1]{\adjustbox{height=1.25cm,center}{#1}}
    \caption{Hyperparameter analysis of our \textsc{AdaFlash} using Qwen3-8B as target model on GSM8K.}
    \label{tab:hyperparameter_analysis}
    \vspace{-1mm}
    \begin{minipage}[t]{0.32\textwidth}
        \textbf{(a)}~Analysis of the mixing coefficient $\alpha$\\[1.8mm]
        \centering
        \hyperparamtab{
            \begin{tabular}{l c c c}
                \toprule
                $\alpha$ & $\tau$         & TPS             & Acc. Rate       \\
                \midrule
                $0.0$    & 7.633          & 802.90          & 0.442           \\
                $0.2$    & 7.713          & 811.61          & 0.448           \\
                $0.4$    & 7.724          & 823.85          & 0.448           \\
                $0.6$    & 7.732          & 825.04          & 0.449           \\
                $0.8$    & \textbf{7.751} & \textbf{835.32} & \textbf{0.450}  \\
                $1.0$    & 7.734          & 828.85          & 0.449           \\
                \bottomrule
            \end{tabular}
        }
        \label{tab:ablation_alpha}
    \end{minipage}
    \hfill
    \begin{minipage}[t]{0.32\textwidth}
        \textbf{(b)}~Analysis of the div. clipping threshold $\delta$\\[1mm]
        \centering
        \hyperparamtab{
            \begin{tabular}{l c c c}
                \toprule
                $\delta$ & $\tau$ & TPS & Acc. Rate \\
                \midrule
                No Clip  & 7.684 & 808.01 & 0.446 \\
                $0.003$  & 7.722 & 739.41 & 0.448 \\
                $0.005$  & 7.725 & 754.73 & 0.448 \\
                $0.010$   & \textbf{7.751} & \textbf{835.32} & \textbf{0.450} \\
                $0.015$  & 7.740 & 824.34 & 0.449 \\
                $0.020$   & 7.734 & 815.76 & 0.449 \\
                \bottomrule
            \end{tabular}
        }
        \label{tab:ablation_clip}
    \end{minipage}
    \hfill \hspace{2mm}
    \begin{minipage}[t]{0.32\textwidth}
        \textbf{(c)}~Analysis of the adaptive length scale factor $\gamma$\\[1mm]
        \centering
        \hyperparamtab{
            \begin{tabular}{l c c c}
                \toprule
                $\gamma$ & Speedup        & $\tau$          & Verify Len.     \\
                \midrule
                $0.8$    & 1.257          & 6.514          & 9.033           \\
                $0.9$    & 1.262          & 6.950          & 10.162          \\
                $1.0$    & \textbf{1.270} & 7.270          & 11.271          \\
                $1.1$    & 1.231          & 7.522          & 12.342          \\
                $1.2$    & 1.202          & 7.618          & 13.117          \\
                $1.3$    & 1.185          & 7.675          & 13.674          \\
                \bottomrule
            \end{tabular}
        }
        \label{tab:gamma_analysis}
    \end{minipage}
    \vspace{-2mm}
\end{table*}

In the upper part of Table~\ref{tab:ablation}, we evaluate the on-policy distillation components in our \textsc{AdaFlash} framework while keeping the adaptive length head enabled. With the standard reverse-KL loss and neither of the two components, the drafter achieves moderate speedup but is limited by the distribution mismatch between the drafter and the target model. Introducing entry-wise divergence clipping consistently improves both the acceptance length and speedup across concurrency levels. Further incorporating the mixture OPD loss brings additional gains by introducing hard-label supervision on the target top-1 token. Combining both components, the full configuration achieves the highest acceptance length and speedup.

In the lower part of Table~\ref{tab:ablation}, we ablate the adaptive length head while keeping the full OPD recipe. Without the length head, we use a fixed verification length of $11$, matching the average verification length ($11.271$) of \textsc{AdaFlash} on GSM8K. In this setting, the speedup at high concurrency ($C{=}128$) drops to $1{\times}$, offering no improvement over standard AR decoding. Adding the length head with fixed parameters substantially restores the high-concurrency speedup, as it truncates the candidate sequence to its viable prefix and eliminates wasteful verification. Enabling online updates of the length head yields further improvement, since the head co-adapts with the evolving drafter, maintaining accurate acceptance-rate predictions throughout deployment. These results confirm that the adaptive length head effectively reduces the verification waste caused by \emph{token-level variance}, and that its online update is essential for sustaining this benefit as the drafter evolves.

\noindent \textbf{Hyperparameter Analysis.~~}
To answer \textbf{Q3}, we investigate the impact of key hyperparameters on the performance of our \textsc{AdaFlash} framework.
We first examine the mixing coefficient $\alpha$, which interpolates between reverse-KL ($\alpha=0$) and hard forward KL ($\alpha=1$). We sweep $\alpha \in \{0.0, 0.2, 0.4, 0.6, 0.8, 1.0\}$ on Qwen3-8B with a concurrency of 1 for (a) and (b), and 128 for (c), and report results in Table~\ref{tab:ablation_alpha}. All three metrics improve monotonically from $\alpha=0.0$ to $\alpha=0.8$, confirming that the hard forward KL term provides effective top-1 supervision that the pure reverse-KL objective lacks. Setting $\alpha=1.0$ causes a slight regression, indicating that the mode-seeking regularization from reverse-KL remains beneficial even when hard-label supervision dominates.

We further ablate the entry-wise divergence clipping threshold $\delta$ and report results in Table~\ref{tab:ablation_clip}. Most thresholds improve $\tau$ and acceptance rate over the no-clipping baseline, confirming that divergence clipping suppresses noisy large-magnitude gradients. However, overly aggressive clipping ($\delta\leq0.005$) reduces TPS and acceptance length, suggesting it over-constrains the gradient and slows convergence. The moderate threshold $\delta=0.01$ achieves the best balance.
We also investigate the adaptive length scale factor $\gamma$ and report results in Table~\ref{tab:gamma_analysis}. The speedup is relatively stable across the range $\gamma \in [0.8, 1.3]$, varying between $1.19\times$ and $1.27\times$, indicating that the performance is not overly sensitive to this hyperparameter.
While larger $\gamma$ produces longer draft sequences and thus higher acceptance length $\tau$, the increased verification length offsets the gain, leading to slightly lower overall speedup. The best speedup is achieved at $\gamma=1.0$.

\section{Conclusion}
\label{sec:conclusion}

In this paper, we uncover a fundamental challenge in diffusion-based speculative decoding: the global dependency arising from bidirectional attention and KV injection enables one-pass parallel drafting but also introduces high variance in draft quality, manifesting at both the domain level and the token level. Existing methods rely on fixed, offline-trained diffusion drafters with a constant verification length, which leads to degraded acceptance rates on out-of-distribution domains and wasteful verification of low-quality draft tokens. To this end, we propose \textsc{AdaFlash}, an adaptive framework with two complementary components: (i)~on-policy distillation with a mixture of reverse-KL and hard-label cross-entropy, combined with entry-wise divergence clipping, which continuously adapts the drafter to the deployment distribution; and (ii)~an adaptive length head that predicts the viable draft length in each round and truncates the draft sequence accordingly, substantially lowering the target model's verification cost. We further introduce infrastructure improvements, including an asynchronous training--inference pipeline and adaptive request scheduling. Experiments across eight benchmarks including two long-sequence tasks, and three target models, covering both dense and MoE architectures, demonstrate that \textsc{AdaFlash} consistently outperforms previous SOTA methods, achieving up to $5.3\times$ speedup over AR decoding, with significant gains under high concurrency.

% \section*{Acknowledgements}
% The authors would like to thank XXX

% \newpage
\bibliography{refer}
\bibliographystyle{colm2026_conference}

\newpage
\appendix

\section{Related Work}
\label{sec:related-work}

In this section, we discuss the related topics. 
\vspace{2mm}
\noindent \textbf{Speculative Decoding.~~}
Speculative decoding~\citep{ICML'23:Speculative,arxiv'23:speculative-sampling} accelerates large language model inference by pairing a lightweight draft model with a large target model: the draft model proposes a candidate token sequence, which the target model verifies in parallel, accepting all tokens up to the first rejected one and resampling at the rejection point.

A central line of work focuses on improving the draft sequence's quality.
EAGLE~\citep{ICML'24:EAGLE} introduces feature-level autoregression with token-conditioned drafting.
EAGLE-2~\citep{EMNLP'24:EAGLE-2} extends it with context-aware dynamic draft trees, and EAGLE-3~\citep{arxiv'25:EAGLE-3} further replaces feature prediction with direct token prediction via multi-layer feature fusion.
Lookahead decoding~\citep{ICML'24:Lookahead} avoids a separate draft model entirely by maintaining $n$-gram pools from the Jacobi decoding trajectory for parallel verification.

On the other hand,
SpecTR~\citep{NeurIPS'23:SpecTr} formulates draft selection as an optimal transport problem, while MCSD~\citep{arxiv'24:MCSD} organizes multiple candidate sequences into batches for parallel verification to improve the acceptance rate.
JetSpec~\citep{arxiv'26:JetSpec} trains a causal parallel draft head over fused target-model hidden states, combining one-forward drafting efficiency with branch-wise causal conditioning to achieve longer acceptance lengths.
To adaptively tune hyperparameters during inference, BanditSpec~\citep{ICML'25:BanditSpec} casts the selection as a multi-armed bandit problem, and HedgeSpec~\citep{arxiv'25:Not-a-Bandit} introduces an auxiliary verification step that evaluates all draft models without additional target model queries.
Beyond token-level parallelism, Lookahead Reasoning~\citep{arxiv'25:lookaheadR} exploits step-level parallelism by having the draft model propose entire reasoning steps, which are then expanded in parallel and verified semantically.

\vspace{2mm}
\noindent \textbf{Diffusion Language Models.~~}
Diffusion language models (dLLMs) have emerged as a promising alternative to autoregressive generation, and the release of open-source foundation dLLMs has accelerated research in this area. LLaDA~\citep{arxiv'25:LLaDA} is a native 8B dLLM trained from scratch; Dream~\citep{arxiv'25:Dream} is a 7B model initialized from a pretrained AR checkpoint; and LLaDA 2.0~\citep{technical-report-llada2} scales to 100B parameters via conversion from AR models.
Subsequent efforts improve dLLMs in both efficiency and performance~\citep{arxiv'25:Efficient-DLM,arxiv'25:D2F,arxiv'25:dParallel,arxiv'26:Dmax}. Fast-dLLM~\citep{arxiv'25:Fast-dllm} introduces a block-wise approximate KV-cache to accelerate dLLM inference, and dKV~\citep{NeurIPS'25:dKV} proposes a delayed KV-cache strategy tailored for the non-autoregressive setting~\citep{arxiv'25:SSD,arxiv'25:dInfer,arxiv'25:CTRL}. d3LLM~\citep{arxiv'26:d3llm} improves the efficiency of dLLMs by leveraging distillation. On the other hand, MMaDA~\citep{NeurIPS'25:MMaDA} applies policy-gradient reinforcement learning to enhance dLLM performance across modalities; ReFusion~\citep{arxiv'25:ReFusion} adopts slot-level parallel decoding initialized from Qwen-3-8B; and TraDo~\citep{arxiv'25:Trado} incorporates trajectory-aware reinforcement learning for reasoning tasks.

Several recent works explore dLLMs as drafters for speculative decoding. SpecDiff~\citep{NAACL'25:SpecDiffusion} first demonstrated the feasibility of parallel diffusion drafting, showing that a discrete diffusion model with few denoising steps can match the wall-clock speedup of AR drafters by amortizing drafting cost across positions. DiffuSpec~\citep{ACL'26:DiffuSpec} extended this idea by employing pretrained large-scale dLLMs as drafters. SpecDiff-2~\citep{MLSys'26:SpecDiff2} then mitigates the draft--verifier mismatch between AR and dLLM via streak-distillation (optimizing expected acceptance length) and test-time self-selection (choosing the best draft from multiple low-cost samples). Building on this line, DFlash~\citep{ICML'26:DFlash} proposed a lightweight block diffusion drafter, and conditions each drafting layer on fused multi-layer hidden features from the target model, achieving both low drafting cost and high acceptance rates.
Our work uncovers a previously overlooked pitfall of diffusion drafters: the global dependency arising from bidirectional attention and KV injection, while enabling one-pass parallel generation, also introduces \emph{high variance} in draft quality across both domains and token positions, which limits the achievable speedup. Based on this insight, we propose \textsc{AdaFlash} to address this issue through online drafter adaptation and adaptive verification length.

\vspace{2mm}
\noindent \textbf{On-policy Distillation.~~}
On-policy distillation~\citep{ICLR'24:OPD,ICLR'24:MiniLLM,tml'25:OPD,ICLR'25:SpecKD} trains a student model on sequences sampled from its own policy rather than from a fixed dataset or the teacher, thereby eliminating the train–inference distribution mismatch (exposure bias) that arises when a model is only supervised on offline data. Since the supervision signals are derived from the student's on-policy generations, the student directly corrects the errors it actually encounters during inference, yielding more effective adaptation than offline distillation.

This paradigm also broadens the space of supervision signals. Standard SFT minimizes the cross-entropy against the dataset's ground-truth hard labels, which corresponds to a forward KL objective, and classical knowledge distillation~\citep{TKDE'04:NeC4.5,arxiv'15:KD} similarly matches the teacher's output. In contrast, by operating on student-sampled sequences, on-policy distillation admits more flexible divergences such as the reverse-KL and generalized objectives, as exemplified by GKD~\citep{ICLR'24:OPD}, which mitigate the mode-covering pathologies of forward KL and better suit the student's limited capacity.

On-policy distillation is naturally suited for training the draft model in speculative decoding, as the draft model must approximate the target distribution along trajectories it generates itself. DistillSpec~\citep{ICLR'24:DistillSpec} shows that the token-level acceptance rate equals $1-D_{\mathrm{TV}}(p,q)$ and systematically compares FKL, RKL, JSD, and TVD under different data sources, finding that on-policy data is crucial while the best divergence depends on the task and decoding strategy. In this work, we treat the divergence as a design choice and adopt a mixture of hard-label cross-entropy and clipped reverse-KL tailored to the high-entropy outputs of diffusion drafters. OSD~\citep{ICML'24:OSD} applies continuous distillation from the target model to the draft model on live query distributions, updating the draft based on discrepancies identified during the verification stage. Extending this perspective, OnlineSPEC~\citep{ICML'26:OnlineSPEC} formulates the on-policy update of the draft model as an online learning problem and establishes regret bounds for the evolving draft, thus providing a unified theoretical framework for online speculative decoding. Aurora~\citep{arxiv'26:SpecTrain} further integrates speculator training and deployment into a single closed-loop system.
Draft-OPD~\citep{arxiv'26:Draft-OPD} applies on-policy distillation to speculative drafters including DFlash as an offline post-training stage: it collects target-assisted rollouts on a fixed prompt pool, replays drafting from the verification-exposed error positions, and applies forward KL on accepted tokens and position-decayed reverse KL on rejected ones. In contrast, our OPD continuously updates the drafter on live deployment trajectories.

\vspace{2mm}
\noindent \textbf{Dynamic Candidate Length.~~}
Adaptive candidate length has been studied in the context of autoregressive speculative decoding.
SpecDec++~\citep{COLM'25:SpecDecPP} formulates the candidate length selection as a Markov decision process, trains an acceptance prediction head on top of the AR draft model to predict per-token conditional acceptance probabilities, and stops the current speculation round when the cumulative rejection probability exceeds a threshold.
AdaEAGLE~\citep{arxiv'24:AdaEAGLE} uses a lightweight MLP to regress the draft length of the AR drafter for the next round.
In the context of diffusion-based drafting, DiffuSpec~\citep{ACL'26:DiffuSpec} introduces an adaptive draft-length controller that adjusts the next round's proposal size based on exponential moving averages of recent acceptance and generation lengths.
FailFast~\citep{arxiv'25:FailFastWinBig} takes a different strategy: it dynamically extends the speculation length when the drafter's token-level confidence exceeds a threshold.
Unlike these approaches, our adaptive length head targets diffusion drafters: it directly predicts the acceptance length of the draft sequence to truncate the verification length, and is continuously updated online to co-adapt with the evolving drafter, whereas the predictors of SpecDec++ and AdaEAGLE are trained offline and kept fixed. We also note that FailFast's strategy of dynamically extending the speculation length is orthogonal and complementary to our approach of truncating the verification length.

\newpage
\section{Additional Experiments}
\label{sec:additional_experiments}
In this section, we provide additional experimental results.

\subsection{Implementation Details}
\label{sec:implementation_details}

\noindent \textbf{Metrics.~~} We evaluate all methods using the following metrics: \rom{1} \emph{Speedup}, defined as the ratio of tokens per second (TPS) of each method to that of standard AR decoding, measuring wall-clock efficiency. \rom{2} \emph{Average accepted length} ($\tau$), i.e., how many tokens of the draft sequence are accepted in one speculative round on average, measuring the quality of the draft model.

\noindent \textbf{Implementation Details.~~} We use Qwen3-8B, Qwen3-Coder-30B-A3B~\citep{qwen3}, and Qwen3.5-9B~\citep{qwen3.5} as target models, covering dense and mixture-of-experts architectures of different scales. For the DFlash drafter, we use a candidate length of $k=15$ (i.e., block size $= 16$) per speculative round. Reported verification lengths include the anchor token.
The on-policy distillation uses a mixture loss with mixing coefficient $\alpha = 0.8$, reverse-KL temperature $= 1.0$, and entry-wise divergence clipping threshold $\delta = 0.01$.
The drafter is updated with AdamW at learning rate $3 \times 10^{-4}$, while the adaptive length head is trained separately at $2 \times 10^{-4}$ with MSE loss.
In our AdaFlash configuration, the adaptive length head is disabled when the concurrency $\leq 16$, in which case the candidate length is fixed at $k=15$: the adaptive length head is only activated under higher concurrency scenarios to dynamically reduce candidate lengths and lower verification cost.
Online training is triggered whenever the replay buffer accumulates $128$ on-policy samples; each buffer batch is trained for $2$ epochs with batch size $1$ and gradient accumulation steps of $2$ (effective batch size $2$), up to a maximum sequence length of $2{,}048$ tokens.
The online adaptation runs in a streaming fashion via an asynchronous training--inference pipeline: an inference worker serves incoming queries on dedicated GPUs while a training worker updates the drafter in the background on separate GPUs.
The system is built on top of SGLang~\citep{NeurIPS'24:sglang} for high-throughput serving with continuous batching.
We employ mixed-precision training with bfloat16 and use Flash Attention~\citep{NeurIPS'22:FlashAttention} to accelerate attention computation. Greedy decoding (temperature $=0$) is adopted during inference across all experiments unless explicitly stated otherwise.

\subsection{Additional Experimental Results}
\label{sec:additional_experiments:additional_results}

\begin{table}[t]
    \centering
    \caption{Ablation study validating the high variance issue in diffusion drafters. All drafters are trained from scratch using Qwen3-4B as the target model with $800$K mixed-domain samples for $2$ epochs. DFlash variants use block size $k=16$; EAGLE-3 uses tree size $60$.}
    \vspace{-2mm}
    \label{tab:variance_ablation}
    \resizebox{0.85\textwidth}{!}{
        \begin{tabular}{l *{3}{cc} cc}
            \toprule
            \multirow{2}{*}{Method} & \multicolumn{2}{c}{HumanEval} & \multicolumn{2}{c}{GSM8K} & \multicolumn{2}{c}{MT-Bench} & \multicolumn{2}{c}{Cross-Domain} \\
            \cmidrule(lr){2-3} \cmidrule(lr){4-5} \cmidrule(lr){6-7} \cmidrule(lr){8-9}
                                    & $\tau$ & Speedup              & $\tau$ & Speedup            & $\tau$ & Speedup              & Mean($\tau$) & Std($\tau$)            \\
            \midrule
            DFlash                  & 4.59   & 3.15$\times$         & 5.34   & 3.67$\times$       & 2.69   & 1.88$\times$         & 4.21         & 1.115                  \\
            DFlash (causal)         & 4.41   & 3.02$\times$         & 5.15   & 3.47$\times$       & 2.64   & 1.81$\times$         & 4.07         & 1.053                  \\
            DFlash (AR-like)        & 5.10   & 1.05$\times$         & 5.64   & 1.19$\times$       & 2.67   & 0.56$\times$         & 4.47         & 1.292                  \\
            EAGLE-3                 & 3.34   & 1.95$\times$         & 3.70   & 2.14$\times$       & 3.16   & 1.82$\times$         & 3.40         & 0.225                  \\
            \bottomrule
        \end{tabular}
    }
    \vspace{3mm}
\end{table}

\begin{figure}[t]
    \centering
    \includegraphics[width=\linewidth]{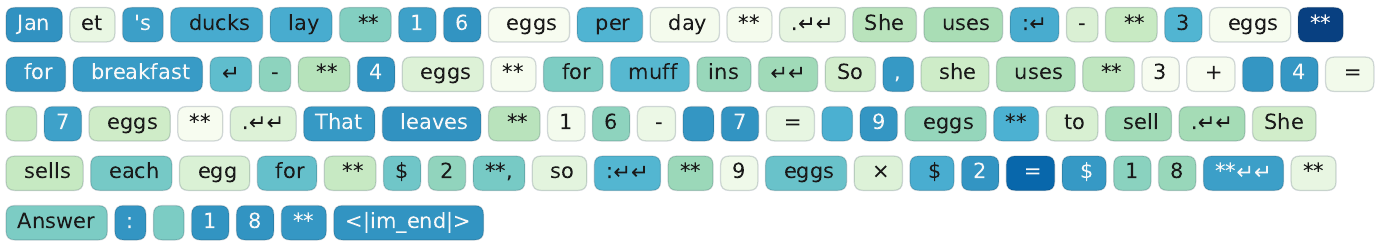}
    \vspace{-5mm}
    \caption{Reverse-KL divergence between the draft and target distributions on a GSM8K response. Each cell is a generated token, shaded by its divergence (darker indicates higher divergence).}
    \label{fig:div_clipping_example}
\end{figure}
\noindent \textbf{Validating the Variance Issue in Diffusion Drafters.~~}
To validate the high variance issue identified in Section~\ref{sec:observation}, we ablate the key components of DFlash by comparing three configurations trained from scratch on Qwen3-4B ($800$K mixed-domain samples, $2$ epochs): \rom{1}~DFlash (bidirectional attention, one-pass parallel generation, KV injection); \rom{2}~DFlash~(causal), replacing bidirectional with causal attention; and \rom{3}~EAGLE-3, an AR drafter that reuses hidden states without KV injection. DFlash variants use candidate length $k=15$; EAGLE-3 uses tree size $60$.

As shown in Table~\ref{tab:variance_ablation}, removing bidirectional attention (DFlash causal) reduces $\mathrm{Std}(\tau)$ from $1.12$ to $1.05$. EAGLE-3, which removes both bidirectional attention and KV injection, achieves the lowest variance ($\mathrm{Std}(\tau)=0.23$), but at the cost of notably lower acceptance lengths ($\tau = 3.40$ vs.\ $4.21$). These results suggest that bidirectional attention and KV injection, i.e., the two components of global dependency, are both sources of high variance in diffusion drafters, yet they are also key to the strong drafting capability of DFlash. This also indicates that the \emph{primary contributor to DFlash's high acceptance rate is KV injection, rather than the diffusion-based architecture itself.}

\vspace{2mm}
\noindent \textbf{Entry-wise divergence clipping.~~}
Unlike AR drafters, diffusion drafters rely on bidirectional attention to predict all draft tokens jointly in a single forward pass, which tends to spread probability mass broadly across the vocabulary.
As a result, the reverse-KL at each draft position, which is a sum of per-entry terms over the vocabulary, is dominated by a few entries on which the drafter places non-negligible mass but the target does not. Consequently, as shown in Figure~\ref{fig:div_clipping_example}, the per-token divergence is highly uneven across the sequence, with a few tokens reaching values orders of magnitude larger than the rest.
During on-policy distillation, such excessive high-divergence entries dominate the gradient and can mislead the overall parameter update, destabilizing training.
Entry-wise divergence clipping addresses this by bounding each entry's KL contribution at a threshold $\delta$ before summation, suppressing outlier gradients while preserving the informative updates from the remaining entries.

\noindent \textbf{Results of \textsc{AdaFlash} on Qwen3.5-9B.~~}
Table~\ref{tab:performance_qwen35} reports results on Qwen3.5-9B, which adopts a hybrid Gated DeltaNet architecture. The \emph{MTP} (Multi-Token Prediction)~\citep{arxiv'24:DeepSeek-V3} method in the table refers to the native multi-token prediction capability of Qwen3.5.
At $C=1$, \textsc{AdaFlash} achieves the highest average speedup of $3.58\times$ across all six benchmarks, outperforming DFlash, OSD, and the MTP head.
At higher concurrency, \textsc{AdaFlash} consistently leads on the math and code benchmarks, although its gains on ShareGPT narrow or reverse slightly when $C \geq 32$.
We attribute this to an engineering limitation: SGLang's current support for the Gated DeltaNet architecture does not yet allow a fully efficient implementation of variable-length verification scheduling, so the adaptive length head incurs additional overhead under high-concurrency serving on this model.
We expect this high-concurrency gap to close as SGLang's support for Gated DeltaNet matures.

\begin{table*}[t]
    \vspace{1mm}
    \centering
    \caption{Performance Comparison on Qwen3.5-9B}
    \vspace{-3mm}
    \label{tab:performance_qwen35}
    \resizebox{\textwidth}{!}{
        \begin{tabular}{lcc *{7}{cc}}
            \toprule
            \multirow{3}{*}{Method}            & \multirow{3}{*}{Conc.} & \multicolumn{4}{c}{\textsc{Math}} & \multicolumn{4}{c}{\textsc{Code}} & \multicolumn{2}{c}{\textsc{Chat}}    & \multicolumn{2}{c}{\textsc{Hybrid}} & \multicolumn{2}{c}{}                                                                                                                                                                                     \\
            \cmidrule(lr){3-6} \cmidrule(lr){7-10} \cmidrule(lr){11-12} \cmidrule(lr){13-14}
                                   &                        & \multicolumn{2}{c}{MathQA}        & \multicolumn{2}{c}{GSM8K}         & \multicolumn{2}{c}{OpenCodeInstruct} & \multicolumn{2}{c}{CodeAlpaca}      & \multicolumn{2}{c}{ShareGPT} & \multicolumn{2}{c}{Blend} & \multicolumn{2}{c}{\textit{Avg.}}                                                                                                             \\
            \cmidrule(lr){3-16}
                                   &                        & Speedup                           & $\tau$                            & Speedup                              & $\tau$                              & Speedup                      & $\tau$                    & Speedup                           & $\tau$ & Speedup               & $\tau$ & Speedup               & $\tau$ & Speedup               & $\tau$ \\
            \midrule
            \multirow{4}{*}{MTP} & 1                      & 2.24$\times$ & 3.54 & 2.08$\times$ & 3.54 & 2.17$\times$ & 3.48 & 1.88$\times$ & 3.23 & 1.81$\times$ & 2.77 & 2.10$\times$ & 3.38 & 2.05$\times$ & 3.32 \\
                                   & 32                     & 1.64$\times$ & 3.54 & 1.43$\times$ & 3.54 & 1.51$\times$ & 3.48 & 1.35$\times$ & 3.24 & \textbf{1.42$\times$} & 2.77 & \textbf{1.53$\times$} & 3.38 & 1.48$\times$ & 3.32 \\
                                   & 64                     & 1.29$\times$ & 3.54 & 1.19$\times$ & 3.54 & 1.29$\times$ & 3.48 & 1.16$\times$ & 3.23 & \textbf{1.13$\times$} & 2.77 & \textbf{1.25$\times$} & 3.37 & 1.22$\times$ & 3.32 \\
                                   & 128                    & 0.98$\times$ & 3.54 & 0.92$\times$ & 3.54 & 0.98$\times$ & 3.49 & 0.88$\times$ & 3.23 & \textbf{0.85$\times$} & 2.77 & 0.95$\times$ & 3.38 & 0.93$\times$ & 3.32 \\
            \cmidrule(l){1-16}
            \multirow{4}{*}{DFlash} & 1                      & 3.69$\times$ & 6.92 & 3.59$\times$ & 6.54 & 3.79$\times$ & 7.77 & 2.87$\times$ & 5.20 & 1.91$\times$ & 3.33 & 2.96$\times$ & 6.20 & 3.14$\times$ & 5.99 \\
                                   & 32                     & 1.53$\times$ & 6.92 & 1.43$\times$ & 6.54 & 1.57$\times$ & 7.77 & 1.28$\times$ & 5.21 & 0.90$\times$ & 3.34 & 1.27$\times$ & 6.23 & 1.33$\times$ & 6.00 \\
                                   & 64                     & 1.08$\times$ & 6.92 & 1.07$\times$ & 6.53 & 1.20$\times$ & 7.77 & 0.97$\times$ & 5.20 & 0.62$\times$ & 3.34 & 0.92$\times$ & 6.20 & 0.98$\times$ & 5.99 \\
                                   & 128                    & 0.81$\times$ & 6.92 & 0.85$\times$ & 6.53 & 0.93$\times$ & 7.77 & 0.73$\times$ & 5.20 & 0.47$\times$ & 3.33 & 0.70$\times$ & 6.22 & 0.75$\times$ & 6.00 \\
            \cmidrule(l){1-16}
            \multirow{4}{*}{OSD} & 1                      & 4.26$\times$ & 8.62 & 4.30$\times$ & 8.28 & 4.31$\times$ & 9.35 & 3.11$\times$ & 6.11 & 1.97$\times$ & 3.68 & 3.14$\times$ & 7.02 & 3.52$\times$ & 7.18 \\
                                   & 32                     & 1.72$\times$ & 8.63 & 1.67$\times$ & 8.28 & 1.72$\times$ & 9.31 & 1.34$\times$ & 6.10 & 0.92$\times$ & 3.65 & 1.33$\times$ & 7.04 & 1.45$\times$ & 7.17 \\
                                   & 64                     & 1.20$\times$ & 8.62 & 1.27$\times$ & 8.27 & 1.34$\times$ & 9.33 & 1.02$\times$ & 6.09 & 0.63$\times$ & 3.62 & 0.96$\times$ & 7.05 & 1.07$\times$ & 7.16 \\
                                   & 128                    & 0.91$\times$ & 8.63 & 1.01$\times$ & 8.27 & 1.02$\times$ & 9.33 & 0.79$\times$ & 6.11 & 0.47$\times$ & 3.63 & 0.74$\times$ & 7.04 & 0.82$\times$ & 7.17 \\
            \cmidrule(l){1-16}
            \multirow{4}{*}{\textsc{AdaFlash}} & 1                      & \textbf{4.33$\times$} & 8.88 & \textbf{4.32$\times$} & 8.42 & \textbf{4.39$\times$} & 9.63 & \textbf{3.24$\times$} & 6.39 & \textbf{1.99$\times$} & 3.88 & \textbf{3.23$\times$} & 7.25 & \textbf{3.58$\times$} & 7.41 \\
                                   & 32                     & \textbf{1.87$\times$} & 7.87 & \textbf{1.76$\times$} & 7.55 & \textbf{1.78$\times$} & 8.29 & \textbf{1.51$\times$} & 5.94 & 1.08$\times$ & 3.70 & 1.49$\times$ & 6.60 & \textbf{1.58$\times$} & 6.66 \\
                                   & 64                     & \textbf{1.44$\times$} & 6.61 & \textbf{1.41$\times$} & 6.71 & \textbf{1.47$\times$} & 6.97 & \textbf{1.33$\times$} & 4.84 & 1.04$\times$ & 2.98 & 1.23$\times$ & 5.17 & \textbf{1.32$\times$} & 5.55 \\
                                   & 128                    & \textbf{1.16$\times$} & 6.65 & \textbf{1.24$\times$} & 6.72 & \textbf{1.25$\times$} & 7.04 & \textbf{1.10$\times$} & 4.83 & 0.83$\times$ & 2.99 & \textbf{0.99$\times$} & 5.18 & \textbf{1.09$\times$} & 5.57 \\
            \bottomrule
        \end{tabular}
    }
    % \vspace{-5mm}
\end{table*}

\noindent \textbf{Long-Sequence Generation.~~}
To evaluate the effectiveness of \textsc{AdaFlash} on extended reasoning tasks, we conduct experiments on MATH-500 and AIME25 with a maximum output length of $32{,}768$ tokens and thinking mode enabled, using Qwen3-8B as the target model. The results are reported in Tables~\ref{tab:long_sequence_math500} and~\ref{tab:long_sequence_aime25}. On MATH-500, \textsc{AdaFlash} achieves a speedup of $3.67\times$ at $C=1$ and $3.52\times$ at $C=32$, outperforming DFlash ($2.60\times$ and $2.46\times$), OSD ($3.51\times$ and $3.40\times$), and EAGLE-3 ($2.26\times$ and $1.82\times$). On AIME25, \textsc{AdaFlash} achieves $3.18\times$ at $C=1$ and $2.77\times$ at $C=32$. These results indicate that \textsc{AdaFlash} maintains its advantages in long-sequence generation.

% \FloatBarrier

\noindent \textbf{Cross-Domain Generalization.~~}
To evaluate whether offline OPD can produce a drafter that generalizes across domains without online adaptation, we compare three settings on Qwen3-8B: \rom{1} the offline DFlash drafter, \rom{2} \textsc{AdaFlash} with in-domain online OPD (training and evaluation on the same benchmark), and \rom{3} \textsc{AdaFlash} trained offline via OPD on \textsc{PerfectBlend}, a mixed-domain dataset distinct from each test benchmark, and evaluated with fixed weights without further online updates. Table~\ref{tab:cross_domain} reports results on four benchmarks (MathQA, GSM8K, OpenCodeInstruct, and CodeAlpaca) at concurrency $C \in \{1, 32, 64, 128\}$.
In-domain \textsc{AdaFlash} generally achieves the best speedup and acceptance length, confirming the benefit of adapting the drafter to the deployment domain. A minor exception is GSM8K, where the cross-domain drafter slightly outperforms the in-domain one, likely because the limited size of the GSM8K training set restricts the volume of on-policy samples for effective in-domain adaptation. Overall, the cross-domain drafter with fixed weights outperforms the offline DFlash baseline across all settings by leveraging mixed-domain OPD training, yet still lags behind in-domain online adaptation on most benchmarks. This confirms that on-policy updates are necessary to adapt the drafter to the current domain.

\vspace{1mm}
\noindent \textbf{Results on NPU.~~}
We further evaluate DFlash and cross-domain \textsc{AdaFlash} on Huawei Ascend 910C NPU with SGLang-NPU as the serving backend, using Qwen3-8B as the target model.
As shown in Table~\ref{tab:cross_domain_npu}, cross-domain \textsc{AdaFlash} consistently outperforms DFlash across all concurrency levels.
These results demonstrate that \textsc{AdaFlash} achieves similar speedup gains on NPU as on GPU, confirming its generalizability across hardware backends.
In the future, we plan to adapt \textsc{AdaFlash} to the OpenPangu-2.0-Flash model~\citep{misc'26:openPangu-2.0-Flash}.

\begin{table*}[t]
    \centering
    \begin{minipage}[t]{0.48\textwidth}
        \centering
        \caption{Long-sequence generation results on \textsc{MATH-500} dataset (32K output length, Qwen3-8B, thinking mode enabled).}
        \vspace{-1mm}
        \label{tab:long_sequence_math500}
        \resizebox{\textwidth}{!}{
            \begin{tabular}{l c c c c}
                \toprule
                Method                             & Conc. & Speedup               & $\tau$ & Acc. Rate \\
                \midrule
                \multirow{2}{*}{EAGLE-3}           & 1     & 2.26$\times$          & 4.25  & 0.105 \\
                                                   & 32    & 1.82$\times$          & 4.37  & 0.109 \\
                \multirow{2}{*}{DFlash}            & 1     & 2.60$\times$          & 3.97  & 0.198 \\
                                                   & 32    & 2.46$\times$          & 4.04  & 0.203 \\
                \multirow{2}{*}{OSD}               & 1     & 3.51$\times$          & 5.28  & 0.285 \\
                                                   & 32    & 3.40$\times$          & 5.31  & 0.287 \\
                \multirow{2}{*}{\textsc{AdaFlash}} & 1     & \textbf{3.67}$\times$ & \textbf{5.37}  & \textbf{0.291} \\
                                                   & 32    & \textbf{3.52}$\times$ & \textbf{5.14}  & \textbf{0.276} \\
                \bottomrule
            \end{tabular}
        }
    \end{minipage}
    \hfill
    \begin{minipage}[t]{0.48\textwidth}
        \centering
        \caption{Long-sequence generation results on \textsc{AIME25} dataset (32K output length, Qwen3-8B, thinking mode enabled).}
        \vspace{-1mm}
        \label{tab:long_sequence_aime25}
        \resizebox{\textwidth}{!}{
            \begin{tabular}{l c c c c}
                \toprule
                Method                             & Conc. & Speedup               & $\tau$ & Acc. Rate \\
                \midrule
                \multirow{2}{*}{EAGLE-3}           & 1     & 2.30$\times$          & 4.16  & 0.102 \\
                                                   & 32    & 1.68$\times$          & 4.26  & 0.105 \\
                \multirow{2}{*}{DFlash}            & 1     & 1.89$\times$          & 3.32  & 0.155 \\
                                                   & 32    & 1.54$\times$          & 3.36  & 0.157 \\
                \multirow{2}{*}{OSD}               & 1     & 3.01$\times$          & 5.21  & 0.281 \\
                                                   & 32    & 2.52$\times$          & 5.24  & 0.283 \\
                \multirow{2}{*}{\textsc{AdaFlash}} & 1     & \textbf{3.18}$\times$ & \textbf{5.32}  & \textbf{0.288} \\
                                                   & 32    & \textbf{2.77}$\times$ & \textbf{4.98}  & \textbf{0.265} \\
                \bottomrule
            \end{tabular}
        }
    \end{minipage}
\end{table*}

\begin{table*}[t]
    \vspace{-1mm}
    \centering
    \caption{Cross-domain generalization results on Qwen3-8B. We compare offline DFlash with our \textsc{AdaFlash} (trained on \textsc{PerfectBlend}), a mixed-domain dataset distinct from each test benchmark, and evaluated with fixed weights without further online adaptation. In-domain \textsc{AdaFlash} results are reported in Table~\ref{tab:performance_comparison}.}
    \label{tab:cross_domain}
    \resizebox{\textwidth}{!}{
        \begin{tabular}{lc *{4}{cc} cc}
            \toprule
            \multirow{3}{*}{Method} & \multirow{3}{*}{Conc.} & \multicolumn{4}{c}{\textsc{Math}} & \multicolumn{4}{c}{\textsc{Code}} & \multicolumn{2}{c}{}                                                                                                 \\
            \cmidrule(lr){3-6} \cmidrule(lr){7-10}
                                    &                        & \multicolumn{2}{c}{MathQA} & \multicolumn{2}{c}{GSM8K} & \multicolumn{2}{c}{OpenCodeInstruct} & \multicolumn{2}{c}{CodeAlpaca} & \multicolumn{2}{c}{\textit{Avg.}}                                                                                                 \\
            \cmidrule(lr){3-12}
                                    &                        & Speedup                    & $\tau$                    & Speedup                              & $\tau$                         & Speedup                           & $\tau$        & Speedup               & $\tau$        & Speedup               & $\tau$        \\
            \midrule
            \multirow{4}{*}{DFlash}
                                    & 1                      & 4.38$\times$               & 7.09                      & 3.84$\times$                         & 6.27                           & 4.08$\times$                      & 6.69          & 3.39$\times$          & 5.63          & 3.92$\times$          & 6.42          \\
                                    & 32                     & 1.86$\times$               & 7.10                      & 1.59$\times$                         & 6.28                           & 1.77$\times$                      & 6.68          & 1.46$\times$          & 5.66          & 1.67$\times$          & 6.43          \\
                                    & 64                     & 1.24$\times$               & 7.11                      & 1.03$\times$                         & 6.29                           & 1.17$\times$                      & 6.67          & 0.91$\times$          & 5.65          & 1.09$\times$          & 6.43          \\
                                    & 128                    & 0.90$\times$               & 7.11                      & 0.82$\times$                         & 6.28                           & 0.86$\times$                      & 6.67          & 0.74$\times$          & 5.66          & 0.83$\times$          & 6.43          \\
            \cmidrule(l){1-12}
            \multirow{4}{*}{\shortstack[l]{\textsc{AdaFlash}                                                                                                                                                                                                                                                                      \\(cross-domain)}}
                                    & 1                      & 4.88$\times$               & 8.37                      & 4.65$\times$                & 8.10                  & 4.11$\times$                      & 6.98          & 3.43$\times$          & 5.93          & 4.27$\times$          & 7.35          \\
                                    & 32                     & 2.07$\times$               & 8.39                      & 1.84$\times$                & 8.11                  & 1.76$\times$                      & 6.96          & 1.48$\times$          & 5.95          & 1.79$\times$          & 7.35          \\
                                    & 64                     & 1.55$\times$               & 7.91                      & 1.37$\times$                & 7.79                  & 1.38$\times$                      & 6.65          & 1.13$\times$          & 5.53          & 1.36$\times$          & 6.97          \\
                                    & 128                    & 1.29$\times$               & 7.93                      & 1.35$\times$                & 7.76                  & 1.15$\times$                      & 6.64          & 1.07$\times$          & 5.54          & 1.22$\times$          & 6.97          \\
            \bottomrule
        \end{tabular}
    }
    \vspace{-1mm}
\end{table*}

\begin{table*}[t]
    \vspace{-1mm}
    \centering
    \caption{Cross-domain generalization on NPU with Qwen3-8B, using the same offline DFlash and cross-domain \textsc{AdaFlash} checkpoints as in Table~\ref{tab:cross_domain}.}
    \label{tab:cross_domain_npu}
    \resizebox{\textwidth}{!}{
        \begin{tabular}{lc *{4}{cc} cc}
            \toprule
            \multirow{3}{*}{Method} & \multirow{3}{*}{Conc.} & \multicolumn{4}{c}{\textsc{Math}} & \multicolumn{4}{c}{\textsc{Code}} & \multicolumn{2}{c}{}                                                                                                 \\
            \cmidrule(lr){3-6} \cmidrule(lr){7-10}
                                    &                        & \multicolumn{2}{c}{MathQA} & \multicolumn{2}{c}{GSM8K} & \multicolumn{2}{c}{OpenCodeInstruct} & \multicolumn{2}{c}{CodeAlpaca} & \multicolumn{2}{c}{\textit{Avg.}}                                                                                                 \\
            \cmidrule(lr){3-12}
                                    &                        & Speedup                    & $\tau$                    & Speedup                              & $\tau$                         & Speedup                           & $\tau$        & Speedup               & $\tau$        & Speedup               & $\tau$        \\
            \midrule
            \multirow{4}{*}{DFlash}
                                    & 1                      & 3.43$\times$               & 7.10                      & 3.01$\times$                         & 6.28                           & 3.43$\times$                      & 6.69          & 2.63$\times$          & 5.63          & 3.13$\times$          & 6.43          \\
                                    & 32                     & 1.77$\times$               & 7.09                      & 1.41$\times$                         & 6.28                           & 1.48$\times$                      & 6.67          & 1.43$\times$          & 5.68          & 1.52$\times$          & 6.43          \\
                                    & 64                     & 1.42$\times$               & 7.09                      & 1.17$\times$                         & 6.26                           & 1.33$\times$                      & 6.70          & 1.10$\times$          & 5.64          & 1.26$\times$          & 6.42          \\
                                    & 128                    & 1.24$\times$               & 7.12                      & 1.02$\times$                         & 6.27                           & 1.14$\times$                      & 6.69          & 1.03$\times$          & 5.65          & 1.11$\times$          & 6.43          \\
            \cmidrule(l){1-12}
            \multirow{4}{*}{\shortstack[l]{\textsc{AdaFlash} \\(cross-domain)}}
                                    & 1                      & 4.23$\times$               & 8.45                      & 3.99$\times$                         & 8.11                           & 3.57$\times$                      & 6.98          & 2.96$\times$          & 5.93          & 3.69$\times$          & 7.37          \\
                                    & 32                     & 1.90$\times$               & 8.49                      & 1.49$\times$                         & 8.09                           & 1.74$\times$                      & 6.97          & 1.37$\times$          & 5.96          & 1.63$\times$          & 7.38          \\
                                    & 64                     & 1.50$\times$               & 8.00                      & 1.39$\times$                         & 7.79                           & 1.33$\times$                      & 6.65          & 1.10$\times$          & 5.54          & 1.33$\times$          & 7.00          \\
                                    & 128                    & 1.58$\times$               & 8.01                      & 1.61$\times$                         & 7.79                           & 1.47$\times$                      & 6.65          & 1.32$\times$          & 5.53          & 1.50$\times$          & 7.00          \\
            \bottomrule
        \end{tabular}
    }
    \vspace{-1mm}
\end{table*}

\vspace{2mm}
\begin{wraptable}{r}{0.42\textwidth}
    \centering
    \vspace{-4mm}
    \caption{Comparison of the adaptive length head with DSpark's confidence head on GSM8K using Qwen3-8B. The confidence threshold is $0.94$ and $\gamma = 1.0$.}
    \vspace{-1mm}
    \setstretch{1.08}
    \resizebox{0.42\textwidth}{!}{
        \begin{tabular}{l c c c}
            \toprule
            Method              & $\tau$ & Verify Len. & Speedup \\
            \midrule
            \multicolumn{4}{c}{\textit{Concurrency $= 64$}} \\
            DFlash ($k\!=\!15$) & 6.408  & 16.000      & 1.09$\times$ \\
            DSpark's Head       & 5.071  & 6.673       & 1.21$\times$ \\
            Adaptive Len Head   & \textbf{5.164} & \textbf{6.448} & \textbf{1.23}$\times$ \\
            \cmidrule(l){1-4}
            \multicolumn{4}{c}{\textit{Concurrency $= 128$}} \\
            DFlash ($k\!=\!15$) & 6.402  & 16.000      & 0.79$\times$ \\
            DSpark's Head       & 5.074  & 6.673       & 1.01$\times$ \\
            Adaptive Len Head   & \textbf{5.150} & \textbf{6.437} & \textbf{1.03}$\times$ \\
            \bottomrule
        \end{tabular}
    }
    \label{tab:compare_dspark}
    \vspace{-5mm}
\end{wraptable}

\noindent \textbf{Comparison with DSpark.~~}
DSpark~\citep{misc'26:DSpark} employs a per-position confidence head that estimates the acceptance probability at each draft position and truncates the sequence once it drops below a threshold, whereas our adaptive length head directly predicts the expected acceptance length for the entire sequence (Section~\ref{sec:adaptive-length-head}). We compare the two approaches on GSM8K (both using DFlash's backbone, only offline fine-tuning the confidence head/length head) using Qwen3-8B at concurrency $C \in \{64, 128\}$, with results reported in Table~\ref{tab:compare_dspark}. Both methods substantially reduce the verification length from $k+1=16$ to approximately $6$--$7$ tokens, confirming the benefit of dynamic length adjustment under high concurrency. However, our adaptive length head consistently achieves a higher acceptance length with a shorter verification length: at $C=64$, $\tau = 5.164$ vs.\ $5.071$ and verification length $6.448$ vs.\ $6.673$, yielding a speedup of $1.23\times$ compared to $1.21\times$; at $C=128$, $\tau = 5.150$ vs.\ $5.074$ and verification length $6.437$ vs.\ $6.673$. This improvement can be attributed to our formulation: directly predicting the overall expected acceptance length avoids the compounding errors inherent in sequential per-position probability estimation, providing a more accurate estimation of the quality of the draft sequence.

\begin{table*}[t]
    \vspace{3mm}
    \centering
    \caption{Effect of decoding temperature on \textsc{AdaFlash} (Qwen3-8B). We evaluate greedy decoding ($T=0$) and sampling with $T=1$ on four benchmarks at concurrency $C \in \{1, 32\}$.}
    \vspace{-1mm}
    \label{tab:temperature}
    \resizebox{\textwidth}{!}{
        \begin{tabular}{lc *{4}{cc} cc}
            \toprule
            Method & Conc. & \multicolumn{2}{c}{MathQA} & \multicolumn{2}{c}{GSM8K} & \multicolumn{2}{c}{OpenCodeInstruct} & \multicolumn{2}{c}{CodeAlpaca} & \multicolumn{2}{c}{\textit{Avg.}} \\
            \midrule
            \multicolumn{2}{l}{Temperature = 0}       & Speedup                    & $\tau$                    & Speedup                              & $\tau$                         & Speedup                           & $\tau$        & Speedup               & $\tau$        & Speedup               & $\tau$        \\
            \midrule
            \multirow{2}{*}{DFlash}
                                 & 1                      & 4.38$\times$               & 7.09                      & 3.84$\times$                         & 6.27                           & 4.08$\times$                      & 6.69          & 3.39$\times$          & 5.63          & 3.92$\times$          & 6.42          \\
                                 & 32                     & 1.86$\times$               & 7.10                      & 1.59$\times$                         & 6.28                           & 1.77$\times$                      & 6.68          & 1.46$\times$          & 5.66          & 1.67$\times$          & 6.43          \\
            \cmidrule(l){1-12}
            \multirow{2}{*}{AdaFlash}
                                 & 1                      & \textbf{5.32$\times$}               & 9.83                      & \textbf{4.54$\times$}                         & 7.75                           & \textbf{4.87$\times$}                      & 8.40          & \textbf{3.75$\times$}          & 6.80          & \textbf{4.62$\times$}          & 8.20          \\
                                 & 32                     & \textbf{2.27$\times$}               & 9.84                      & \textbf{1.83$\times$}                         & 7.76                           & \textbf{2.03$\times$}                      & 8.39          & \textbf{1.58$\times$}          & 6.79          & \textbf{1.93$\times$}          & 8.20          \\
            \midrule
            \multicolumn{2}{l}{Temperature = 1}       & Speedup                    & $\tau$                    & Speedup                              & $\tau$                         & Speedup                           & $\tau$        & Speedup               & $\tau$        & Speedup               & $\tau$        \\
            \midrule
            \multirow{2}{*}{DFlash}
                                 & 1                      & 3.32$\times$               & 5.85                      & 3.48$\times$                         & 5.70                           & 3.59$\times$                      & 5.82          & 2.97$\times$          & 5.03          & 3.34$\times$          & 5.60          \\
                                 & 32                     & 1.52$\times$               & 5.84                      & 1.38$\times$                         & 5.74                           & 1.58$\times$                      & 5.82          & 1.32$\times$          & 5.03          & 1.45$\times$          & 5.61          \\
            \cmidrule(l){1-12}
            \multirow{2}{*}{AdaFlash}
                                 & 1                      & \textbf{3.56$\times$}               & 6.75                      & \textbf{3.80$\times$}                         & 6.69                           & \textbf{3.85$\times$}                      & 6.67          & \textbf{3.05$\times$}          & 5.61          & \textbf{3.56$\times$}          & 6.43          \\
                                 & 32                     & \textbf{1.62$\times$}               & 6.74                      & \textbf{1.45$\times$}                         & 6.73                           & \textbf{1.66$\times$}                      & 6.58          & \textbf{1.34$\times$}          & 5.62          & \textbf{1.52$\times$}          & 6.42          \\
            \bottomrule
        \end{tabular}
    }
    % \vspace{-1mm}
\end{table*}

\noindent \textbf{Different Temperature.~~}
We evaluate \textsc{AdaFlash} under greedy decoding ($T=0$) and stochastic sampling ($T=1$) on four benchmarks at $C \in \{1, 32\}$ with Qwen3-8B. Table~\ref{tab:temperature} reports the results.
As expected, sampling reduces the average acceptance length compared to greedy decoding ($\tau \approx 6.43$ vs.\ $8.20$ at $C=1$), since the target model's output distribution becomes more diffuse under $T=1$ and can reduce their distributional overlap, thereby lowering the acceptance probability.
Nevertheless, \textsc{AdaFlash} retains a substantial speedup under sampling: $3.56\times$ at $C=1$ and $1.52\times$ at $C=32$ on average, demonstrating that the on-policy distillation and adaptive length head remain effective across decoding regimes.

\vspace{2mm}
\begin{wraptable}{r}{0.42\textwidth}
    \centering
    \vspace{-4mm}
    \caption{From-scratch training using Qwen3-1.7B as target model on GSM8K dataset. The number in parentheses denotes the draft budget.}
    \vspace{-1mm}
    \resizebox{0.42\textwidth}{!}{
        \begin{tabular}{l c c c}
            \toprule
            Method                 & Avg. TPS       & $\tau$        & Speedup               \\
            \midrule
            AR Decoding            & 455.86          & 1.000          & 1.00$\times$          \\
            EAGLE-3 (16)           & 611.58          & 3.586          & 1.34$\times$          \\
            \textsc{AdaFlash} (16) & \textbf{839.55} & \textbf{4.554} & \textbf{1.84}$\times$ \\
            \bottomrule
        \end{tabular}
    }
    \vspace{-2mm}
    \label{tab:from_scratch}
\end{wraptable}

\noindent \textbf{Training From Scratch.~~}
Beyond online adaptation of a pre-trained drafter, \textsc{AdaFlash} can also train a diffusion drafter entirely from scratch, which is useful when no off-the-shelf DFlash drafter is available for a given target model. To demonstrate this, we use Qwen3-1.7B as the target model, train \textsc{AdaFlash} from scratch on the GSM8K training split ($7.4$K samples) for $10$ epochs, and evaluate it on the GSM8K test split ($1.3$K samples).
As shown in Table~\ref{tab:from_scratch}, the from-scratch drafter trained with \textsc{AdaFlash} achieves an average acceptance length of $4.554$ and a $1.84\times$ speedup over AR decoding, substantially outperforming EAGLE-3 (using its publicly released pre-trained weights from~\citet{AngelSlim2025}), which attains an acceptance length of $3.586$ and a $1.34\times$ speedup under the same setting. These results confirm that \textsc{AdaFlash} is effective not only for online adaptation but also for training diffusion drafters from scratch.

\end{document}